\documentclass{article}

\usepackage[preprint]{neurips_2024}

\usepackage[utf8]{inputenc} 
\usepackage[T1]{fontenc}    
\usepackage{url}            
\usepackage{booktabs}       
\usepackage{amsfonts}       
\usepackage{nicefrac}       
\usepackage{microtype}      
\usepackage{algorithm}
\usepackage[noend]{algpseudocode}
\usepackage[table]{xcolor}
\usepackage{amssymb,amsmath,amsthm, amsfonts}

 \usepackage{mathrsfs}
\usepackage{caption}
\usepackage{capt-of}
\usepackage{lipsum}
\usepackage{multibib}
\newcites{app}{References}
\usepackage{txfonts}
\usepackage[colorlinks]{hyperref}

\usepackage[toc,page,header]{appendix}

\usepackage{multirow}

\newcommand{\R}[0]{\mathbb{R}}

\newcommand{\head}[1]{\vspace{1.7mm}\noindent{{\textcolor{c1}{\bf #1.}}}}

\usepackage{nicefrac}       
\usepackage{microtype}      
\usepackage{xcolor}         

\definecolor{c1}{HTML}{765D97} 
\definecolor{c2}{HTML}{7BA67D}
\definecolor{c3}{HTML}{fc6160}
\definecolor{myblue}{HTML}{E6F3FC} 
\definecolor{mygray}{HTML}{DBE2E9} 
\definecolor{mygreen}{HTML}{006400} 

\hypersetup{
    colorlinks=true,
    linkcolor=black,
    urlcolor=c1,
    citecolor=c2,
}

\newcommand{\squishlist}{
 \begin{list}{$\bullet$}
  { \setlength{\itemsep}{0pt}
     \setlength{\parsep}{3pt}
     \setlength{\topsep}{3pt}
     \setlength{\partopsep}{0pt}
     \setlength{\leftmargin}{1.5em}
     \setlength{\labelwidth}{1em}
     \setlength{\labelsep}{0.5em} } }

\newcommand{\squishlisttwo}{
 \begin{list}{}
  { \setlength{\itemsep}{0pt}
    \setlength{\parsep}{0pt}
    \setlength{\partopsep}{0pt}
    \setlength{\leftmargin}{2em}
    \setlength{\labelwidth}{1.5em}
    \setlength{\labelsep}{0.5em} } }

\newcommand{\squishend}{
  \end{list}  }

\usepackage{tikz}
\newcommand*\circled[1]{\tikz[baseline=(char.base)]{
            \node[shape=circle,draw,inner sep=0.3pt, fill=c2] (char) {\textcolor{white}{#1}};}}

\title{MambaMixer: Efficient Selective State Space Models with Dual Token and Channel Selection}

\author{%
  Ali Behrouz \\
  Cornell University\\
  \texttt{ab2947@cornell.edu} \\
  \And
    Michele Santacatterina \\
  NYU Grossman School of Medicine\\
  \texttt{santam13@nyu.edu} \\
  \newline
  \\
  \\
  \newline
  \href{https://MambaMixer.github.io}{Project Page (Code \& Models)}
  \And
    Ramin Zabih \\
  Cornell University\\
  \texttt{rdz@cs.cornell.edu} \\
}

\usepackage{minitoc}

\begin{document}

\maketitle

\doparttoc 
\faketableofcontents 

\begin{abstract}
  Recent advances in deep learning have mainly relied on Transformers due to their data dependency and ability to learn at scale. The attention module in these architectures, however, exhibit quadratic time and space in input size, limiting their scalability for long-sequence modeling. Despite recent attempts to design efficient and effective architecture backbone for multi-dimensional data, such as images and multivariate time series, existing models are either data independent, or fail to allow inter- and intra-dimension communication. Recently, State Space Models (SSMs), and more specifically Selective State Space Models (S6), with efficient hardware-aware implementation, have shown promising potential for long sequence modeling. Motivated by the recent success of SSMs, we present MambaMixer block, a new architecture with data dependent weights that uses a dual selection mechanism across tokens and channels–called Selective Token and Channel Mixer. MambaMixer further connects the sequential selective mixers using a weighted averaging mechanism, allowing layers to have direct access to different layers' input and output. As a proof of concept, we design Vision MambaMixer (ViM2) and Time Series MambaMixer (TSM2) architectures based on MambaMixer block and explore their performance in various vision and time series forecasting tasks. Our results underline the importance of selectively mixing across both tokens and channels. In ImageNet classification, object detection, and semantic segmentation tasks, ViM2 achieves competitive performance with well-established vision models, i.e., ViT, MLP-Mixer, ConvMixer, and outperforms SSM-based vision models, i.e., ViM and VMamba. In time series forecasting, TSM2, an attention and MLP-free architecture, achieves outstanding performance compared to state-of-the-art methods while demonstrating significantly improved computational cost. These results show that while Transformers, cross-channel attention, and cross-channel MLPs are sufficient for good performance in practice, neither is necessary.
\end{abstract}

\section{Introduction}
In recent years, Transformers \citep{vaswani2017attention} have been the pivotal backbone architecture behind deep learning’s success, enabling a number of breakthrough advances in language modeling~\citep{wolf2019huggingface}, vision~\citep{dosovitskiy2021ViT}, time series~\citep{zhou2021informer}, healthcare~\citep{Tang2023Brain}, and several other domains~\citep{radford2023robust, behrouz2023unsupervised}. The attention modules in Transformers are crucial for their data dependency and enable them to generalize to unseen data and tasks given the context as input. They, however, are difficult to scale efficiently to long sequences due to their quadratic time and space complexity. Breaking this quadratic computational cost is a key step towards new possibilities for deep learning such as long-range context learning~\citep{gu2023mamba}, large object detection~\citep{zhu2024ViM}, and long-range time series forecasting~\citep{liu2021pyraformer}.

To alleviate this computational complexity bottleneck, several recent studies have focused on designing sub-quadratic sequence models motivated by a diverse range of objectives: i.e., MLP-Mixer~\citep{tolstikhin2021mlp} and  ConvMixer~\citep{trockman2023patches} are motivated as simpler alternatives to attention modules, MonarchMixer (M2)~\citep{fu2023monarch} tackles efficiency without losing quality by using sparse Monarch matrices, and efficient attentions \citep{xiao2024efficient, kacham2023polysketchformer, ding2023longnet, chen2021scatterbrain} sparsify or approximate the full attention matrix. These methods, however, either \circled{1} are based on data-independent parameters,  \circled{2} introduce a trade-off between expressivity and speed, underperforming relative to Transformers when are scalable and efficient, or \circled{3} are actually slow in practice, due to low hardware utilization~\citep{dao2022monarch, chen2021scatterbrain}.

Recently, structured State Space Models (SSMs) have emerged as a promising class of architectures for sequence modeling~\citep{gu2022efficiently, fu2023h3, smith2023convolutional}. SSMs can be seen as a combination of Recurrent Neural Networks (RNNs) and Convolutional Neural Networks (CNNs), making them very efficient  in training (as a CNN) and inference (as an RNN)~\citep{gu2020hippo}. Despite their efficiency, primary general SSM architectures, e.g., S4~\citep{gu2022efficiently}, S4D~\citep{gu2022S4D}, are based on data-independent parameters, limiting their effectiveness in compressing context into a smaller state~\citep{gu2023mamba}. To alleviate this limitation, recently, \citet{gu2023mamba} present Mamba, a selective SSMs (S6) that effectively selects relevant context by making the SSM weights time variant (i.e., data dependent). Mamba achieves on par performance with Transformer-based state-of-the-art methods in language modeling while having less parameters and scaling near-linearly in sequence length~\citep{gu2023mamba}, addressing all the three abovementioned limitations. The success of Mamba motivates several studies to adapt its design to different domains and modalities, e.g., vision~\citep{zhu2024ViM, liu2024vmamba}, graphs~\citep{behrouz2024graph}, videos~\citep{li2024mamband}, DNA modeling~\citep{schiff2024caduceus}, etc.

Surprisingly, Mamba and its variants independently apply the S6 block to each channel, overlooking information flow across channels (also known as \emph{Channel Mixing}). The lack of channel mixing in Mamba not only results in stability issues while scaling to large-sized networks~\citep{patro2024simba}, but it also cause missing the relationships among feature maps, limiting Mamba's ability to model global information in multi-dimensional data such as images and multivariate time series. Using additional channel mixing blocks for S6 blocks, however, might be challenging in large-scale networks as due to their recurrent nature, increasing the number of blocks for both token and channel mixing can damage the information flow, limiting their ability to use early features.

In this paper, we present MambaMixer, an efficient selective state space models with dual selection across both channels and tokens. MambaMixer sequentially uses \emph{Selective Token Mixer} and \emph{Selective Channel Mixer}, each of which consists of a bidirectional S6 block~\citep{gu2023mamba}, to efficiently select and mix (resp. filter) informative (resp. irrelevant) tokens and channels. Contrary to original Mamba block that uses simple skip connections between consecutive blocks, inspired by DenseNet~\citep{huang2017densely} and DenseFormer~\citep{pagliardini2024denseformer}, MambaMixer block allows layers to have direct access to earlier features (i.e., inputs and outputs of earlier layers) and further enhances the information flow between selective channel and token mixer blocks as well as different layers using a weighted averaging mechanism, enabling MambaMixer-based models to use a large number of layers and making training more stable for large networks.

As a proof of concept, we employ MambaMixer block to design \textbf{Vi}sion \textbf{M}amba\textbf{M}ixer (ViM2) and \textbf{T}ime \textbf{S}eries \textbf{M}amba\textbf{M}ixer (TSM2) for vision and time series forecasting tasks, respectively. ViM2 first tokenizes the images and uses Cross-Scan Module (CSM)~\citep{liu2024vmamba} to vertically and horizontally scan images, determining the order of tokens. It then, uses a MambaMixer block to selectively mixes tokens and channels. While TSM2 shares very similar architecture with ViM2, Contrary to ViM2, it uses unidirectional S6 blocks for time dimension due to its causal nature and inspired by \citep{chen2023tsmixer}, it uses an additional MambaMixer block to selectively mix auxiliary information of time series (whenever is available) as well as a 2-dimensional normalization across both time and variate dimensions. We further explore the performance of proposed models in various vision and time series forecasting tasks. In ImageNet classification, object detection, and semantic segmentation tasks, ViM2 achieves competitive performance with well-established vision models, and outperforms SSM-based vision models. In time series forecasting, TSM2 achieves outstanding performance compared to state-of-the-art methods on various datasets with diverse domains.

\head{Contributions} To summarize, our contributions are: \circled{1} Presenting MambaMixer block, a new SSM-based architecture with dual selection that efficiently and effectively selects and mixes (resp. filters) informative (resp. irrelevant) tokens and channels in a data-dependent manner, allowing connections across both channel and token dimensions. \circled{2} Demonstrating the ability and effectiveness of bidirectional S6 blocks to focus on or ignore particular channels using ablation studies. \circled{3} Enhancing information flow in multi-layer MambaMixer-based architectures by allowing direct access to earlier features by a weighted averaging mechanism. \circled{4} Presenting ViM2 and TSM2 models based on MambaMixer for vision and time series forecasting tasks, with outstanding performance compared to baselines.

\begin{table*}
    \centering
    \caption{Comparison of the architecture design of MambaMixer and existing models.}
    \resizebox{\linewidth}{!}{\begin{tabular}{l cc  cc c c}
        \toprule
         \multirow{2}{*}{\textbf{Backbone}}& \multicolumn{2}{c}{\textbf{Token Mixing}} & \multicolumn{2}{c}{\textbf{Channel Mixing}} & & \multirow{2}{*}{\textbf{Models}} \\
         \cmidrule(l){2-3} \cmidrule(l){4-5}
         & \textbf{Data-dependent} & \textbf{Module} & \textbf{Data-dependent}  & \textbf{Module} & \textbf{Complexity} & \\
         \midrule
         \midrule
         \multirow{2}{*}{\textbf{MLP-Mixer}} & \multirow{2}{*}{\textcolor{red}{--}} & \multirow{2}{*}{MLP} & \multirow{2}{*}{\textcolor{red}{--}} & \multirow{2}{*}{MLP} & \multirow{2}{*}{$\mathcal{O}(L^2)$} & \multicolumn{1}{l}{MLP-Mixer~\citep{tolstikhin2021mlp}}\\
         & & & & & & \multicolumn{1}{l}{DynaMixer~\citep{wang2022dynamixer}}\\
         \midrule
         \textbf{ConvMixer} & \textcolor{red}{--} & Conv & \textcolor{red}{--} & Conv & {$\mathcal{O}(L)$} & ConvMixer~\citep{trockman2023patches} \\
         \midrule
        \multirow{2}{*}{\textbf{Convolution}} & \multirow{2}{*}{\textcolor{red}{--}} & \multirow{2}{*}{ConvNet} & \multirow{2}{*}{\textcolor{red}{--}} & \multirow{2}{*}{MLP} & \multirow{2}{*}{$\mathcal{O}(L)$} & \multicolumn{1}{l}{AlexNet~\citep{krizhevsky2012imagenet}}\\
         & & & & & & \multicolumn{1}{l}{ResNet~\citep{He2016resnet}}\\
         \midrule
         \multirow{3}{*}{\textbf{Transformers}} & \multirow{2}{*}{\textcolor{mygreen}{\checkmark}} & \multirow{2}{*}{Attention} & \multirow{2}{*}{\textcolor{red}{--}} & \multirow{2}{*}{MLP} & \multirow{3}{*}{$\mathcal{O}(L^2)$}& \multicolumn{1}{l}{ViT~\citep{dosovitskiy2021ViT}}\\
         & & & & & & \multicolumn{1}{l}{DeIT~\citep{touvron2021training}}\\
         & \multirow{1}{*}{\textcolor{mygreen}{\checkmark}} & \multirow{1}{*}{Attention + Conv} & \multirow{1}{*}{\textcolor{red}{--}} & \multirow{1}{*}{MLP} & & \multicolumn{1}{l}{SwinTransformer~\citep{liu2021swin}}\\
         \midrule
         \multirow{5}{*}{\textbf{SSMs}} & \multirow{2}{*}{\textcolor{red}{--}} & \multirow{2}{*}{SSM} & \multirow{2}{*}{\textcolor{red}{--}} & \multirow{2}{*}{\textcolor{red}{--}} & \multirow{5}{*}{$\mathcal{O}(L)$} & \multicolumn{1}{l}{Hyena~\citep{poli2023hyena}}\\
         &  &  &  &  & & \multicolumn{1}{l}{H3~\citep{fu2023h3}}\\
         & \multirow{2}{*}{\textcolor{mygreen}{\checkmark}} & \multirow{2}{*}{Selective SSM} & \multirow{2}{*}{\textcolor{red}{--}} & \multirow{2}{*}{\textcolor{red}{--}} & & \multicolumn{1}{l}{Mamba~\citep{gu2023mamba}}\\
         & & & & & & \multicolumn{1}{l}{Vim~\citep{zhu2024ViM}}\\
         & \multirow{1}{*}{\textcolor{mygreen}{\checkmark}} & \multirow{1}{*}{Selective SSM} & \multirow{1}{*}{\textcolor{mygreen}{\checkmark}} & \multirow{1}{*}{Selective SSM} & & \multicolumn{1}{l}{MambaMixer~(\textcolor{c2}{Ours})}\\
         \toprule
    \end{tabular}
    }
    \label{tab:channel-mixing-comparison}
\end{table*}

\section{Related Work}
To situate our contributions in a broader context, we discuss related studies in three groups:

\subsection{Sequence Modeling}
Transformers~\citep{vaswani2017attention} have been the pivotal backbone architecture behind deep learning’s success. Despite their outstanding success in various domains, their attention module has quadratic space and time complexity, which limits their scalability. To this end, recently, several studies aim to design attention-free models with competitive performance to Transformers~\citep{karami2024orchid, de2024griffin}. To this end, State-space Models (SSMs) have recently emerged as a powerful and attention-free tools for modeling long input sequences~\citep{poli2023hyena, fu2023h3}. More specifically, recently, \citet{gu2023mamba} present Mamba, a selective state space model that using input-dependent weights can effectively focus on or ignore some particular tokens. While these sequence models show promising performance on 1D data, they overlook the channel-wise dependencies in sequences. Although this channel-wise dependencies in some tasks on 1D input data like language modeling might not significantly damage the performance, its lackness makes the adaption of these sequence models for multi-dimensional data challenging. Our MambaMixer block uses selective state space models~\citep{gu2023mamba} across both channel and tokens to selectively mix and fuse information across these dimensions.

\subsection{Architectures for Generic Vision Backbone}
\head{Convolutional Neural Networks and Mixer Architectures} CNNs have served as the de-facto standard backbone in computer vision since the AlexNet model~\citep{krizhevsky2012imagenet} outperforms vision models designed based on hand-crafted image features~\citep{pinz2006object}. Several studies have focused on improving the design of CNNs: \citet{He2016resnet} present residual networks using skip connection and \citet{ioffe2015batch} introduce batch normalization, both enabling the design of very deep CNNs. Several studies have focused on using sparse convolutions~\citep{xie2017aggregated, liu2015sparse}, e.g., depth-wise convolutions~\citep{guo2019depthwise}. Using the idea of convolutions in the extreme case of small kernels, \citep{tolstikhin2021mlp} present MLP-Mixer that use MLP across both patch and channel directions to fuse information in both spatial and feature directions. The success of MLP-Mixer motivated several studies to use matrix multiplication with efficient and sparse matrices across spatial and channel directions~\citep{wang2022dynamixer, tang2022sparse, fu2023monarch}. All these architectures, however, are based on data-independent weights, limiting their generalizability and in-context learning ability. 

\head{Vision Transformers} Motivated by the success of Transformers~\citep{vaswani2017attention} in various domains, Vision Transformers (ViT)~\citep{dosovitskiy2021ViT} are designed purely based on Transformers. This new backbone architecture has inspired a new paradigm of “isotropic” architectures, which are architectures that use patch embeddings for the first layer and have equal size and shape throughout the network. ViTs due to their ability to learn at scales became popular models over time and various of their variants are designed for a wide array of vision tasks~\citep{liu2021swin, liu2022swin, liu2021image, touvron2021training}. While many ViT architectures use MLPs for channel mixing, very recently, several studies suggest that a simple MLP is not able to effectively filter irrelevant features and discuss the need of channel-wise attentions~\citep{zamir2022restormer, zhang2022resnest}. Despite their outstanding performance, Transformers' time and memory scales quadratic with respect to the input, making utilizing them challenging for high-resolution images. This is even more challenging, specifically for channel-wise attention in large-networks, where the number of channels is large. Our ViM2 model's time scales linearly with respect to the input and shows the same pattern for its memory usage (see Section~\ref{sec:imagenet}). It further uses selective state space models across both channel and token dimension, enabling effectively select (resp. filter) informative (resp. irrelevant) tokens or channels. 

\head{SSM-based Generic Vision Models}
S4ND~\citep{nguyen2022s4nd} is pioneer study to employ SSM blocks for visual tasks, handling visual data as continuous signals across 1D, 2D, and 3D domains. Recently, motivated by the success of Mamba~\citep{gu2023mamba}, Vmamba~\citep{liu2024vmamba} and Vim~\citep{zhu2024ViM} adapt Mamba block for generic vision tasks by addressing the directional sensitivity challenge in SSMs based on bi-directional and cross-scan mechanisms. Subsequently, several studies suggest more sophisticated scan mechanisms to improve the performance of these models~\citep{li2024mamband, huang2024localmamba, hu2024zigma}. Surprisingly, existing adaptions mostly focus on different scanning strategies and treat each channel separately, missing cross-channel dependency. Our ViM2 model, using MambaMixer blocks, can effectively and selectively fuse information across both dimensions.


\subsection{Architectures for Generic Timeseries Backbone}
Not surprisingly, Transformer-based models have been common choices for time series forecasting, when modeling the complex relationships of covariates are required~\citep{zhou2021informer, liu2021pyraformer, wu2021autoformer, ilbert2024unlocking, nie2023a}. Several studies have focused on making the design of Transformers for time series more efficient~\citep{zhou2021informer, wu2021autoformer}. Some other studies have focused on extracting long-term information for better forecasting~\citep{nie2023a, zhou2022film}. To this end, \citet{zhou2022film} and \citet{zhou2022fedformer} suggest decomposing the sequences using Fast Fourier Transformation, enabling better extraction of long-term information. Recently, \citet{chen2023tsmixer} present TSMixer, an all-MLP architecture for time series forecasting, with state-of-the-art performance and show that to achieve competitive performance in practice, Transformers are not necessarily. Our TSM2, is an alternative MLP- and attention-free architecture that is designed based on MambaMixer block. In this design, S6 blocks, in both directions of variates and time, is used to select (resp. filter) important (resp. irrelevant) variates as well as time stamps.

\section{Model: MambaMixer}
In this section, we first discuss preliminary concepts about SSMs and then introduce MambaMixer block in details.

\subsection{Preliminaries}
SSMs are known as linear time-invariant systems that map input sequence $x(t) \in \R^L$ to response sequence $y(t) \in \R^L$~\citep{aoki2013state}. To this end, SSMs use a latent state $h(t) \in \R^{N \times L}$, parameter $\mathbf{A}\in \R^{N\times N}$, and projection parameters $\mathbf{B} \in \R^{N\times 1}, \mathbf{C}\in \R^{1 \times N}$ such that:
\begin{align}\nonumber
    &h'(t) = \mathbf{A} \: h(t) + \mathbf{B} \: x(t), \\
    & y(t) = \mathbf{C} \: h(t).
\end{align}
The above differential equation in deep learning settings, however, is hard to solve and so discrete space state models~\citep{gu2020hippo, zhang2023effectively} suggests discretizing the above system using a parameter $\boldsymbol{\Delta}$, i.e.,
\begin{align}\nonumber
    &h_t = \bar{\mathbf{A}} \: h_{t-1} + \bar{\mathbf{B}} \: x_t,\\ \label{eq:ssm1}
    &y_t = \mathbf{C} \: h_t, 
\end{align}
where 
\begin{align}\nonumber
    &\bar{\mathbf{A}} = \exp\left( \boldsymbol{\Delta} \mathbf{A} \right),\\ \label{eq:ssm_disc}
    &\bar{\mathbf{B}} = \left( \boldsymbol{\Delta} \mathbf{A} \right)^{-1} \left( \exp \left( \boldsymbol{\Delta} \mathbf{A} - I \right) \right) \: . \: \boldsymbol{\Delta} \mathbf{B}.
\end{align}
Discrete SSMs can be interpreted as a combination of CNNs and RNNs and are equivalent to the following convolution~\citep{gu2020hippo}:
\begin{align}\nonumber
    &\bar{\mathbf{K}} = \left( {\mathbf{C}} \bar{\mathbf{B}}, {\mathbf{C}} \bar{\mathbf{A}} \bar{\mathbf{B}}, \dots, {\mathbf{C}} \bar{\mathbf{A}}^{L-1} \bar{\mathbf{B}} \right),\\ \label{eq:ssm_conv}
    &y = x \ast \bar{\mathbf{K}},
\end{align}
which makes them very efficient in both training, as a CNN, and inference, as an RNN.

Despite discrete SSMs efficiency, they are based on data-independent parameters, meaning that parameters $\bar{\mathbf{A}}$, $\bar{\mathbf{B}}$, and ${\mathbf{C}}$ are time invariant and the same for any input, limiting their effectiveness in compressing context into a smaller state~\citep{gu2023mamba}. To alleviate this limitation, recently, \citet{gu2023mamba} present Mamba, a selective SSMs (S6) that effectively selects relevant context by enabling dependence of the parameters $\bar{\mathbf{B}}$, $\bar{\mathbf{C}}$, and $\boldsymbol{\Delta}$ on the input $x_t$, i.e.,:
\begin{equation}
    \bar{\mathbf{B}}_t = \texttt{Linear}_{\textbf{B}}(x_t), \qquad \bar{\mathbf{C}}_t = \texttt{Linear}_{\textbf{C}}(x_t), \qquad \text{and} \quad \boldsymbol{\Delta}_t = \texttt{Softplus}\left(\texttt{Linear}_{\boldsymbol{\Delta}}(x_t)\right),
\end{equation}
where $\texttt{Linear}(.)$ is a linear projection and $\texttt{Softplus}(.) = \log(1 + \exp(.))$. This selection (or data dependency) comes at the cost of the model not being able to be trained as a CNN (\autoref{eq:ssm_conv}), causing challenges for the model efficiency and scalability. To overcome this challenge, \citep{gu2023mamba} show that the linear recurrence in \autoref{eq:ssm1} can be formulated as an associative scan~\citep{martin2018parallelizing}, which accepts efficient parallel algorithms, resulting in logarithmic complexity in sequence length.

\begin{figure*}
    \centering
    \includegraphics[width=\linewidth]{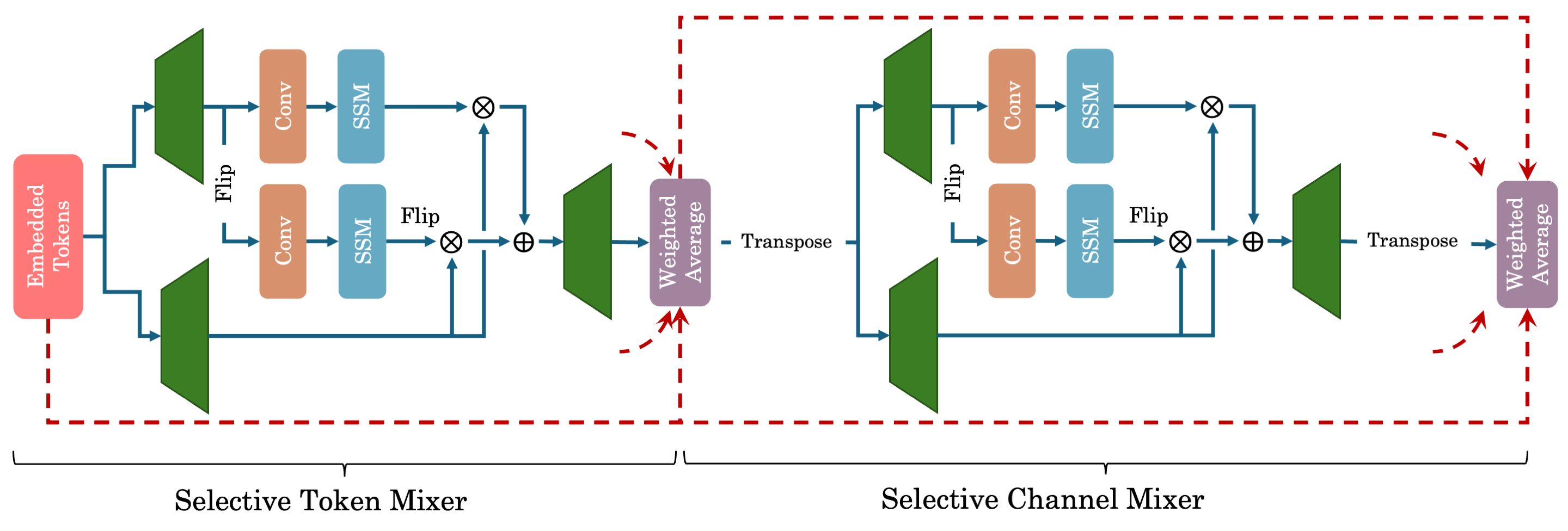}
    \caption{Architecture design of MambaMixer. For further potential architectures see~\autoref{app:arch}.}
    \label{fig:M2Arch}
\end{figure*}

\subsection{MambaMixer}\label{sec:MambaMixer}
As discussed earlier, despite the success of S6 in language modeling~\citep{gu2023mamba}, employing them for non-causal and/or multi-dimensional data is challenging. Despite recent attempts to adapt S6 block and Mamba model to non-causal and/or multi-dimensional data~\citep{li2024mamband, zhu2024ViM}, surprisingly, existing adaptions mostly focus on different scanning strategies and treat each channel separately, missing cross-channel dependency. To overcome this challenge and allow S6 block to select cross channels and tokens, we present MambaMixer block which has three main modules (See~\autoref{fig:M2Arch}): \emph{Selective Token Mixer}, \emph{Selective Channel Mixer}, and \emph{Weighted Averaging of Earlier Features}.

\head{Selective Token Mixer}
The first module of MambaMixer is selective token mixer, which aim to mix and fuse information across tokens while having the ability of focus on or ignore particular tokens. Let $x \in \R^{\texttt{B} \times \texttt{L} \times \texttt{D}}$ represents the input, where \texttt{B} is the batch size, \texttt{L} is the sequence length and \texttt{D} is the number of channels. Inspired by Mamba architecture, we use a S6 block with gated MLP as well as a convolution layer. While Mamba is designed for 1D data and so uses a 1D convolution, we later discuss how the choice of convolution should depend on the input data, i.e., depth-wise convolution~\citep{chollet2017xception} for images and 1D convolution for time series. Therefore, unidirectional Selective Token Mixer module can be summarized as follows:
\begin{align}
    &\bar{\mathbf{x}} = \sigma\left( \texttt{Conv}\left( \texttt{Linear}\left( x \right) \right) \right), \\
    &\bar{\mathbf{x}}_{\texttt{MLP}} = \texttt{MLP}\left( x \right), \\
    &\bar{\mathbf{B}}_x = \texttt{Linear}_{\textbf{B}}(x), \\
    &\bar{\mathbf{C}}_x = \texttt{Linear}_{\textbf{C}}(x) \\
    &\boldsymbol{\Delta}_x = \texttt{Softplus}\left(\texttt{Linear}_{\boldsymbol{\Delta}}(x)\right)\\
    &\mathbf{y}_{\texttt{Token}} = \texttt{SSM}_{\bar{\mathbf{A}}, \bar{\mathbf{B}}_{x}, {\mathbf{C}_x}, \boldsymbol{\Delta}_x}\left( \bar{\mathbf{x}} \right) \otimes \bar{\mathbf{x}}_{\texttt{MLP}} \qquad \qquad \qquad \qquad \qquad \qquad \qquad \:\: \quad \text{\textcolor{c2}{(Using Equation~\ref{eq:ssm1})}}, 
\end{align}
where $\texttt{Conv}(.)$ is a convolution and $\otimes$ is non-linearity. The special case of selective token mixer module can be seen as the Mamba block~\citep{gu2023mamba}, when we use 1D convolution as $\texttt{Conv}(.)$. This design allows the model to not only mix and fuse information across tokens, but also focus on or ignore particular tokens in this process. For multi-dimensional data (e.g., images, videos, etc.), however, tokens are not causal and requires a multi-directional mixing process with multiple scans to allow information flow between each pair of tokens. In these cases, with $d$ possible scanning, we simply extend the above block to $d$-directional selective token mixer by using $d$ unidirectional S6 blocks (instead of one) each of which getting the input sequence based on different scanning, and then added their corresponding outputs together. We further discuss this in Section~\ref{sec:Vim2}, when we adapt MambaMixer for vision tasks.

\head{Selective Channel Mixer}
In multi-dimensional data, learning the dependencies of features is an important step toward training an effective and robust model. For example, it is known that images in different domains share cross-channel correlations while having domain-specific spatial correlations~\citep{guo2019depthwise}. In processing high-resolution images, to avoid high computational cost, instead of explicitly modeling pairwise pixel interactions, one can fuse information across feature channels~\citep{zamir2022restormer}. In multivariate time series, features are different variates and to take advantage of complementary information provided by different variates, learning their dependencies is required. 

Learning the dependencies of features (channel mixing), however, comes with several major challenges: It \circled{1} requires additional parameters, potentially limiting model's scalability, \circled{2} needs to be selective and data-dependent to select (resp. filter) informative (resp. non-informative) features, and \circled{3} requires additional attention to training stability and access to early features, due to additional model depth. To overcome \circled{1} and \circled{2} we use selective channel mixer block that selectively focus on or ignore some particular features, depending on the downstream tasks. In the next part, we further discuss how to address \circled{3}.     

The architecture design is similar to the selective token mixer module with some slight differences. First, due to the non-causal nature of features, we use a bidirectional method to enhance information flow between each pair of features. Second, since this module is a sequence model and is applied across channel dimension, we use \texttt{1D-Conv} as the convolution module. That is, 

\begin{align}
    &\tilde{\mathbf{x}}_{\texttt{forward}} = \sigma\left( \texttt{1D-Conv}\left( \texttt{Linear}\left( x^{\top} \right) \right) \right), \\
    &\tilde{\mathbf{B}}_{x^{\top}} = \texttt{Linear}_{\textbf{B}}(x^{\top}), \qquad \bar{\mathbf{C}}_{x^{\top}} = \texttt{Linear}_{\textbf{C}}(x^{\top}), \qquad \text{and} \quad \boldsymbol{\Delta}_{x^{\top}} = \texttt{Softplus}\left(\texttt{Linear}_{\boldsymbol{\Delta}}(x^{\top})\right),\\
    &z = \texttt{Flip}\left( x \right),\\
    &\tilde{\mathbf{x}}_{\texttt{backward}} = \sigma\left( \texttt{1D-Conv}\left( \texttt{Linear}\left( z^{\top} \right) \right) \right), \\
    &\tilde{\mathbf{B}}_{z^{\top}} = \texttt{Linear}_{\textbf{B}}(z^{\top}), \qquad \bar{\mathbf{C}}_{z^{\top}} = \texttt{Linear}_{\textbf{C}}(z^{\top}), \qquad \text{and} \quad \boldsymbol{\Delta}_{z^{\top}} = \texttt{Softplus}\left(\texttt{Linear}_{\boldsymbol{\Delta}}(z^{\top})\right),\\
    &\tilde{\mathbf{x}}_{\texttt{MLP}} = \texttt{MLP}\left( x^{\top} \right), \\
    &\mathbf{y}_{\texttt{Channel}} = \texttt{SSM}_{\bar{\mathbf{A}}, \bar{\mathbf{B}}_{x^{\top}}, {\mathbf{C}_{x^{\top}}}, \boldsymbol{\Delta}_{x^{\top}}}\left( \tilde{\mathbf{x}}_{\texttt{forward}} \right) \otimes \tilde{\mathbf{x}}_{\texttt{MLP}} + \texttt{Flip}\left(\texttt{SSM}_{\bar{\mathbf{A}}, \bar{\mathbf{B}}_{z^{\top}}, {\mathbf{C}_{z^{\top}}}, \boldsymbol{\Delta}_{z^{\top}}}\left( \tilde{\mathbf{x}}_{\texttt{backward}} \right)\right) \otimes \tilde{\mathbf{x}}_{\texttt{MLP}}. 
\end{align}

This design is similar to Mamba~\citep{gu2023mamba}, when we make Mamba bidirectional by using two modules that are responsible for forward and backward selective scanning and instead of applying it across tokens, we apply it across channels. This channel mixing enhances information flow between each pair of features and can simulate attention-like functionality, while keeping the complexity linear. For example, in vision tasks, while existing selective SSM-based vision models (e.g., ViM~\citep{zhu2024ViM} and VMamba~\citep{liu2024vmamba}) are able to select informative image patches, they cannot focus on a particular part of a single patch. Using the selective channel mixer by selecting informative features, MambaMixer is able to focus on a particular part of a single patch, allowing more flexibility and enhancing generalizibility.

\head{Weighted Averaging of Earlier Features} Inspired by DenseNet~\citep{huang2017densely} and DenseFormer~\citep{pagliardini2024denseformer}, for all MambaMixer layers as well as Selective Token and Channel Mixer blocks, we directly connect the outputs of earlier blocks to their input using a weighted averaging mechanism. This mechanism allows directly re-using early features, capturing complex dynamics and making training more stable. Given $\ell \in \{ 1, \dots, \mathcal{L}\}$, in $\ell$-th MambaMixer block, let $\mathbf{y}^{(\ell)}_{\texttt{Token}}$ and $\mathbf{y}^{(\ell)}_{\texttt{Channel}}$ be the output of selective token and channel mixer blocks, respectively. We compute the input of $\ell$-th selective token mixer block, $x_{\texttt{Token}}^{(\ell)}$ as:
\begin{align}
    {x}_{\texttt{Token}}^{(\ell)} =  \sum_{i = 0}^{\ell - 1} \alpha_{\ell, i} \: \mathbf{y}^{(i)}_{\texttt{Token}} + \sum_{i = 0}^{\ell - 1} \beta_{\ell, i} \: \mathbf{y}^{(i)}_{\texttt{Channel}}.
\end{align}
Similarly, for the input of $\ell$-th selective channel mixer block, $x_{\texttt{Channel}}^{(\ell)}$, we have: 
\begin{align}
    {x}_{\texttt{Channel}}^{(\ell)} =  \sum_{i = 0}^{\ell} \theta_{\ell, i} \: \mathbf{y}^{(i)}_{\texttt{Token}} + \sum_{i = 0}^{\ell - 1} \gamma_{\ell, i} \: \mathbf{y}^{(i)}_{\texttt{Channel}}.
\end{align}
Note that in the above, $\alpha_{i, j}$, $\beta_{i, j}$, $\theta_{i, j}$, and $\gamma_{i, j}$ are learnable parameters and $\mathbf{y}^{(0)}_{\texttt{Token}} = \mathbf{y}^{(0)}_{\texttt{Token}} = x$, where $x$ is the input of the model.

\subsection{Computational Complexity}
To understand the computational complexity (space and time) of MambaMixer, we discuss each of its selective mixers separately. 

\head{Hardware-aware Implementation}
The main challenge of making the weights of SSMs input-dependent is that the \autoref{eq:ssm_conv} in training is not valid anymore, making the training recurrent and slow. Thanks to the hardware-aware implementation of S6 block by \citet{gu2023mamba}, MambaMixer uses their hardware-aware parallel algorithm in recurrent mode for its S6 blocks in both Selective Channel and Token Mixer blocks. Accordingly, there is no restriction for hardware-efficient parallelization of MambaMixer training.

\head{Time Complexity}
Let \texttt{B} be the batch size, \texttt{L} be the length of sequence, \texttt{D} be the hidden state dimension, \texttt{E} be the expanded state dimension, and \texttt{N} be the channel dimension. Using hardware-aware implementation of S6 block, its time complexity for token mixing is $\mathcal{O}\left( \texttt{B} \texttt{L} \texttt{E} + \texttt{E}\texttt{N} \right)$. Similarly, its time complexity for channel mixing is $\mathcal{O}\left( \texttt{B} \texttt{N} \texttt{E} + \texttt{E}\texttt{L} \right)$. Overall, the time complexity of MambaMixer is $\mathcal{O}\left( \texttt{E} \texttt{B}   (\texttt{N} + \texttt{L}) \right)$, which is linear with respect to both sequence length and its number of channels. 

\head{Space Complexity}
One of the main limitation of Transformers~\citep{vaswani2017attention}, and their attention modules is quadratic space complexity with respect to the sequence length. That is, given $\texttt{E} = 2\texttt{D}$, the space complexity of a self-attention block is $\mathcal{O}\left(\texttt{L}\texttt{D}^2 +  \texttt{L}^2\texttt{D}\right)$. For a pure Mamba block for vision, which misses the dependencies of channels, this complexity is $\mathcal{O}\left( \texttt{L}\texttt{D}\texttt{N}\right)$~\citep{zhu2024ViM}. Our MambaMixer similarly requires $\mathcal{O}\left(2 \texttt{L}\texttt{D}\texttt{N} + \mathcal{L} (2\mathcal{L} + 3)\right)$ space ($\mathcal{L}$ is the number of layers), resulting in a linear space complexity with respect to sequence length and number of channels. Note that $\mathcal{L} (2\mathcal{L} + 3)$ is the required memory for our weighted averaging module. In extreme cases, even using 50 MambaMixer blocks, the overhead memory is negligible.

\section{Vision MambaMixer (ViM2)}\label{sec:Vim2}
While in the previous section we focus on designing MambaMixer as a generic backbone that is able to selectively mix both tokens and channels, in this section, we modify and use it to design a vision model.

\begin{figure*}
    \centering
    \includegraphics[width=\linewidth]{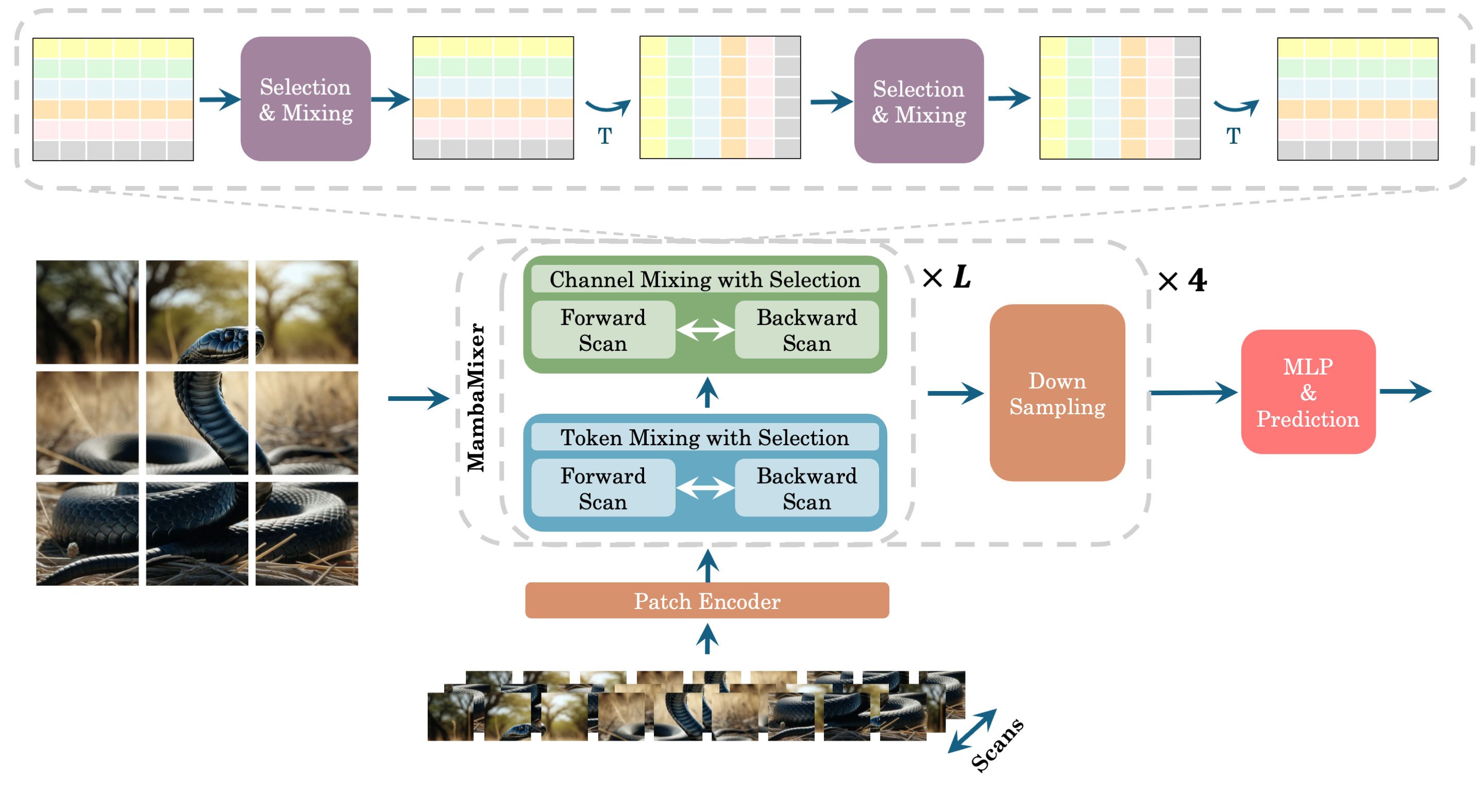}
    \vspace*{-4ex}
    \caption{Architecture design and overview of the ViM2's pipeline.}
    \label{fig:Vim2}
\end{figure*}

\subsection{ViM2 Architecture}
As discussed earlier in Section~\ref{sec:MambaMixer}, The original design of Selective Token Mixer is suitable for causal tokens, while images are non-causal data and requires model modification. In this section, we first describe the modifications of MambaMixer block to adapt it for vision tasks, and then present its overall architecture. 

\head{Scanning}
To adapt the primary design of MambaMixer for images, we use cross-scan presented by~\citet{liu2024vmamba}. In this scan, we scan each image in four directions: top-left to bottom-right, bottom-right to top-left, top-right to bottom-left, and bottom-left to top-right. Then, each scan is passed to a S6 block (selective SSM) and outputs are merged to form new embeddings.

\head{Depth-wise Convolution}
Due to the 2 dimensional nature of images, to preserve the its structure, in Selective Token Mixer, we do not flat patches to 1D sequences. Accordingly, each patch has a 2D shape and so we apply depth-wise convolution~\citep{guo2019depthwise} as the choice of convolution module in the architecture of Selective Token Mixer.

\head{Selective Channel Mixing}
In the selective Channel Mixer, instead of its primary bidirectional design, we use the same scans as Selective Token Mixer, but pass them to the S6 blocks across their channels and finally merged them to form new features.

\head{Overall Architecture}
The overall architecture of ViM2 is illustrated in \autoref{fig:Vim2}. Using a similar pipeline as \citet{liu2024vmamba}, we first, patchify the input images using a stem module, and keep it in the 2D shape without flattening into 1D sequences for token mixing. In the selective channel mixer block, we first flat each token and at the end reshape it to its 2D shape. To learn the hierarchical representation of each image, we then stack multiple modified MambaMixer blocks (described above) with down sampling in four main stages. In fact, each stage (except the first) includes a down sampling in the beginning, resulting in dividing image dimensions by two in each stage, followed by stacks of MambaMixer blocks with weighted averaging modules. We limit these weighted residual connections to each individual stage. Using this design, not only ViM2 can learn hierarchical representations of images, but it also can selectively mix and fuse information across both tokens and channels.

\subsection{Connection to MLP-Mixer and VMamba}
In an extreme case of using SSMs with zero dimension, i.e., removing S6 blocks from both selective token and channel mixer modules, as well as removing depth-wise convolution, ViM2 is equivalent to MLP-Mixer architecture, where $\texttt{MLP}(.)$ is used to fuse information across both channels and tokens. When freezing all $\beta_{i, j} = 0$, $\alpha_{i, j} = 0$ for $j \leq i - 2$, and $\alpha_{i, i-1} = 1$, and/or removing the selective channel mixer block, ViM2 is equivalent to VMamba architecture. Therefore, one can interpret ViM2 as the generalization of VMamba and MLP-Mixer such that it selectively fuses information across both channels and tokens and allows direct access to early features, making the model more robust and stable.

\section{Time Series MambaMixer (TSM2)}
In learning complex temporal patterns in multivariate time series, capturing the dependencies of variates is a key step. To this end, several studies suggest enhancing cross-variate information flow by using a feature mixer module~\citep{chen2023tsmixer, behrouz2023higher}. Existing methods, however, treat all the variates the same in feature mixing phase, including unimportant and/or noisy variates. This becomes a significant challenge when different variates come from different sources and might be noisy (e.g., medical health records), or when some variates are irrelevant to the forecasting task (e.g., traffic forecasting). To overcome this challenge, we adapt our MambaMixer block for multivariate time series forecasting. The Selective Token Mixer (here is also called Selective Time Mixer), can effectively mix and fuse information across time dimension with the ability of focusing on or ignoring specific time stamps. Further, Selective Channel Mixer block (here is also called Selective Variate Mixer) can effectively select (resp. filter) informative (resp. irrelevant) variates. Next, we discuss the details of the TSM2 architecture.

\begin{figure*}
    \centering
    \includegraphics[width=\linewidth]{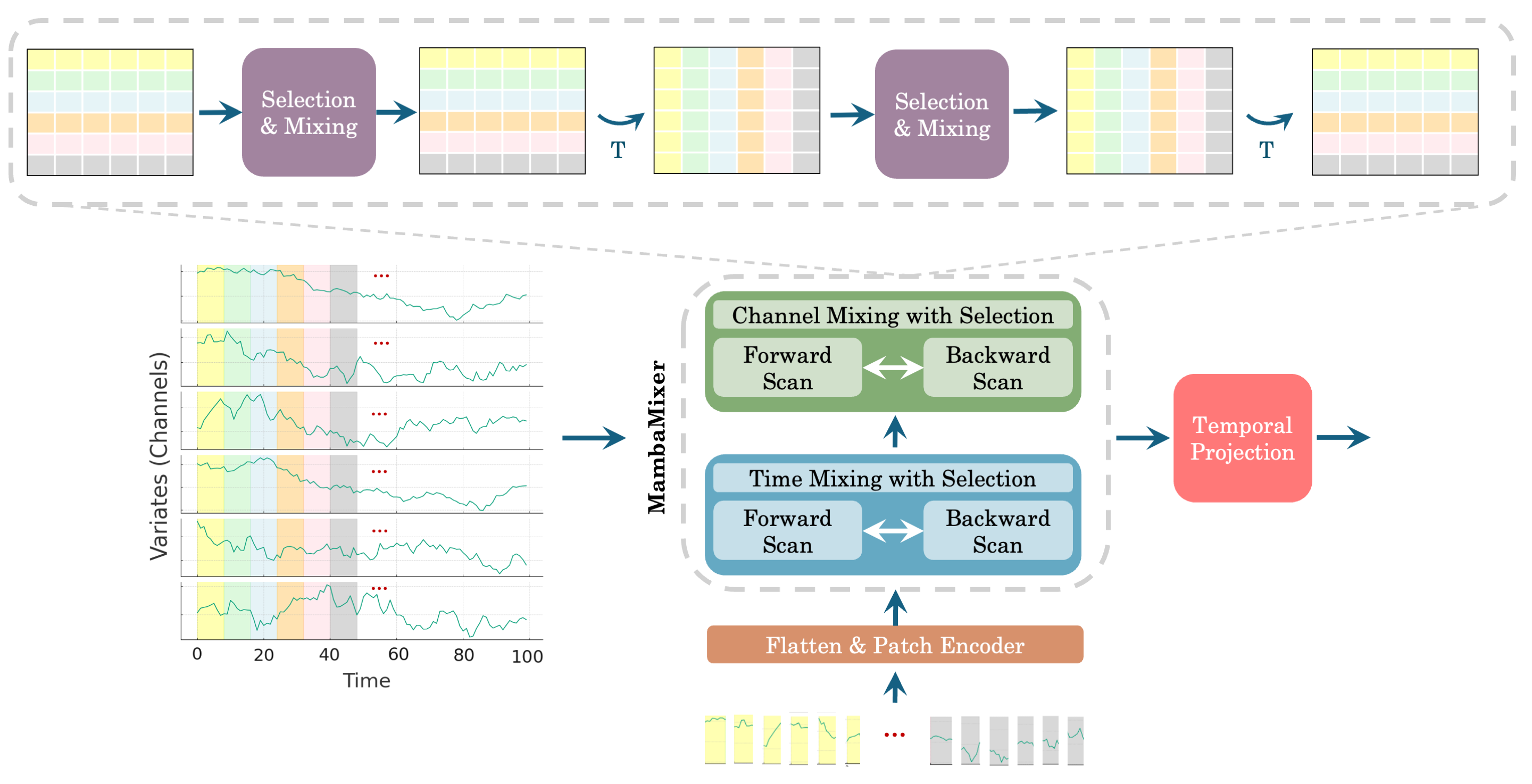}
    \caption{Architecture design and overview of the TSM2's pipeline.}
    \label{fig:TSM2Arch}
\end{figure*}

\subsection{TSM2 Architecture}
The overall architecture of TSM2 is illustrated in \autoref{fig:TSM2Arch}. TSM2 first patchify each variate of the time series separately, and then uses a MambaMixer block with a unidirectional Selective Token (Time) Mixer and a bidirectional Selective Channel (Variate) Mixer to selectively flow information and mix across both time and channels. The main reason for these choices is the causal nature of time dimension and non-causal nature of variates/features. Similar to the primary design of MambaMixer, we also use weighted averaging to allow the model directly access to early features, improving the stability and allowing the model to effectively ignore unnecessary time and channel mixing operations. We apply 2D normalization on both time and feature dimensions and finally, for the choice of convolution in Selective Token (Time) Mixer, we use a 1D convolution, due to the causal nature of the data. 

\head{Auxiliary Information}
\citet{chen2023tsmixer} show the importance of using auxiliary information in time series forecasting, whenever it is available. That is, in addition to the historical observations (multivariate time series), several real-world scenarios provide us with static $\mathbf{S} \in \R^{M \times C_S}$ and future time-varying features $\mathbf{Z} \in \R^{M \times T_Z \times C_Z}$, where $M$ is the number of variates, $T_Z$ is the time dimension of future features, and $C_S$ and $C_Z$ are channel sizes for static and future features, respectively. For example, the static features can be the location of streets in traffic forecasting, the functionality of brain regions in brain activity analysis, meta-data in medical records for survival analysis, etc. Similarly, future time-varying features can be promotion in subsequent weeks, weather forecasting for traffic prediction, hidden symptoms in clinical latency, etc.  

The main challenges to incorporate this information in forecasting tasks is \circled{1} how to align them with the historical observations, and \circled{2} how to effectively encode them to capture their complex dynamic patterns in forecasting tasks. In our TSM2 model, we follow \citet{chen2023tsmixer} pipeline to align historical, future, and static features. That is, given historical multivariate time series data $x \in \R^{M \times T}$ and its corresponding static $\mathbf{S} \in \R^{M \times C_S}$ and future time-varying features $\mathbf{Z} \in \R^{M \times T_Z \times C_Z}$, we first use Selective Channel Mixer on $\mathbf{Z}$ and project it to size $M \times T$, the same size as the historical observation. Similarly, we project $\mathbf{S}$ to the same size and concatenate all the three matrices to use as the input of TSM2. That is, 
\begin{align}
    &\mathbf{S}_{\text{proj}} = \texttt{Linear}_{\text{stat}}(\mathbf{S}) \in \R^{M \times T}, \\
    & \mathbf{Z}_{\text{proj}} = \texttt{Linear}_{\text{future}}\left(\texttt{SelectiveChannelMixer}\left( \mathbf{Z} \right)\right) \in \R^{M \times T},\\
    & x_{\text{input}} = x \:\: || \:\: \mathbf{Z}_{\text{proj}} \:\: || \:\: \mathbf{S}_{\text{proj}} \in \R^{M \times 3T}\\
    & \mathbf{y} = \texttt{TSM2}\left( x_{\text{input}} \right).
\end{align}
Again, due to the causal nature of time-varying auxiliary future features, we use a unidirectional Selective Token Mixer and a bidirectional Selective Channel Mixer. 


\section{Experiments}
In this section, we evaluate the performance of MambaMixer, ViM2, and TSM2 in various tasks of ImageNet classification, object detection, semantic segmentation, and time series forecasting tasks.

\head{ViM2 and TSM2 Models} In our experimental evaluations, we use different variants of our model. While we tune the hyperparameters of TSM2 using grid search to obtain the best performance, for ViM2 we follow the existing studies and use three scales of our model: i.e., ViM2-Tiny, ViM2-Small, and ViM2-Base, referred to as ViM2-T, ViM2-S, and ViM2-B. The details of these architectures are reported in \autoref{tab:scales}. For ablation studies, we further replace our selective channel mixer with MLP without weighted averaging, and refer to it as ViM2-MLP and TSM2-MLP.

\newcommand{\blocka}[3]{\multirow{4}{*}{\(\left[\begin{array}{c}
\text{\texttt{Linear} #1 $\rightarrow 2\times$#1}\\[-.1em]
\text{\texttt{DWConv} 3$\times$3, $2\times$#1}\\[-.1em]
\text{\texttt{S6},  $2\times$#1}\\[-.1em]
\text{\texttt{Linear} 2$\times$#1}\\[-.1em]
\end{array}\right]\)$\times$#2}
}

\newcommand{\blockb}[2]{\multirow{4}{*}{\(\left[\begin{array}{c}
\text{\texttt{CW-Linear} #1}\\[-.1em]
\text{\texttt{CW-S6},  #1}\\[-.1em]
\text{\texttt{CW-Linear} #1 }\\[-.1em]
\end{array}\right]\)$\times$#2}
}

\begin{table}[t]
\centering
\caption{Accuracy comparison across various models on ImageNet-1K$^\dagger$. 
}
\resizebox{.90\linewidth}{!}
{
\begin{tabular}{lccc}
\toprule
Method &  \begin{tabular}[c]{@{}c@{}}Image Size\end{tabular} & \#Parameters (M) & \begin{tabular}[c]{@{}c@{}}ImageNet Top-1 Accuracy\end{tabular} \\
\toprule
\multicolumn{4}{c}{ConvNets}\\
\toprule
ResNet-18~\citep{He2016resnet} & 224$^2$ & 12 & 69.8\\
ResNet-50~\citep{He2016resnet} & 224$^2$ & 25 & 76.2\\
ResNet-101~\citep{He2016resnet}& 224$^2$ & 45 &77.4\\
ResNet-152~\citep{He2016resnet}& 224$^2$ & 60 &78.3\\
\midrule
RegNetY-4G~\citep{radosavovic2020designing} & 224$^2$ & 21 & 80.0 \\
RegNetY-8G~\citep{radosavovic2020designing} & 224$^2$ & 39  & 81.7 \\
RegNetY-16G~\citep{radosavovic2020designing} & 224$^2$ & 84  & 82.9 \\
\midrule
EffNet-B3~\citep{Tan2019EfficientNet} & 300$^2$ & 12  & 81.6 \\
EffNet-B4~\citep{Tan2019EfficientNet} & 380$^2$ & 19 & 82.9 \\
EffNet-B5~\citep{Tan2019EfficientNet} & 456$^2$ & 30 & 83.6 \\
EffNet-B6~\citep{Tan2019EfficientNet} & 528$^2$ & 43   & 84.0 \\
\toprule
\multicolumn{4}{c}{Mixer}\\
\toprule
Mixer-B/16~\citep{tolstikhin2021mlp}& 224$^2$ & 59& 76.4 \\
Mixer-L/16~\citep{tolstikhin2021mlp}&224$^2$ & 207& 71.8\\
\midrule
ConvMixer-768/32~\citep{trockman2023patches} &224$^2$ & 21 & 80.2\\
ConvMixer-1536/20~\citep{trockman2023patches} &224$^2$ & 52 & 81.4\\
\midrule
M2-ViT-b~\citep{fu2023monarch} & 224$^2$ & 45 & 79.5 \\
\toprule
\multicolumn{4}{c}{Transformers}\\
\toprule
ViT-b + Monarch~\citep{fu2023monarch} & 224$^2$ & 33 & 78.9 \\
\midrule
ViT-B/16~\citep{dosovitskiy2021ViT} & 384$^2$ & 86  & 77.9 \\
ViT-L/16~\citep{dosovitskiy2021ViT} & 384$^2$ & 307  & 76.5 \\
\midrule
DeiT-S~\citep{touvron2021training} & 224$^2$ & 22  & 79.8 \\
DeiT-B~\citep{touvron2021training} & 224$^2$ & 86   & 81.8 \\
DeiT-B~\citep{touvron2021training} & 384$^2$ & 86  & 83.1 \\
\midrule
Swin-T~\citep{liu2021swin} & 224$^2$ & 29  & 81.3 \\
Swin-S~\citep{liu2021swin} & 224$^2$ & 50   & 83.0 \\
Swin-B~\citep{liu2021swin} & 224$^2$ & 88   & 83.5 \\
\toprule
\multicolumn{4}{c}{SSMs}\\
\toprule
S4ND-ConvNeXt-T~\citep{nguyen2022s4nd} & 224$^2$ & 30  & 82.2 \\
S4ND-ViT-B~\citep{nguyen2022s4nd} & 224$^2$ & 89  & 80.4 \\
\midrule
VMamba-T~\citep{liu2024vmamba} & 224$^2$ & 22   & 82.2 \\
VMamba-S~\citep{liu2024vmamba} & 224$^2$ & 44  & 83.5 \\
VMamba-B~\citep{liu2024vmamba} & 224$^2$ & 75  & 83.2 \\
\midrule
ViM-T~\citep{zhu2024ViM} & 224$^2$ & 7  & 76.1 \\
ViM-S~\citep{zhu2024ViM} & 224$^2$ & 26  & 80.5 \\
\midrule
\rowcolor{myblue}
ViM2-MLP & 224$^2$ & 40  & 79.1 \\
\midrule
\rowcolor{myblue}
ViM2-T & 224$^2$ & 20   & 82.7 \\
\rowcolor{myblue}
ViM2-S & 224$^2$ & 43   & 83.7 \\
\rowcolor{myblue}
ViM2-B & 224$^2$ & 74  & 83.9 \\
\bottomrule
\multicolumn{4}{l}{$^\dagger$ Reported results are preliminary results and might be changed in next versions.}
\end{tabular}
}
\normalsize
\label{tab:imagenet}
\end{table}

\subsection{Image Classification on ImageNet}\label{sec:imagenet}
\head{Setup}
In this experiment, we evaluate and compare ViM2 classification performance on ImageNet-1K~\citep{deng2009imagenet} with other baselines. Following \citet{liu2024vmamba}, we use the pipeline of \citet{liu2022swin} and train our models for 300 epochs (with the first 20 epochs to warm-up), with batch size of 1024 and using AdamW optimizer~\citep{loshchilov2018decoupled} with a momentum of $0.9$, learning rate of $0.001$, and a weight decay of $0.05$.

\head{Results}
\autoref{tab:imagenet} reports the ImageNet classification results for ViM2 and various baselines based on ConvNets (i.e., ResNet~\citep{He2016resnet}, RegNetY~\citep{radosavovic2020designing}, and EffNet~\citep{Tan2019EfficientNet}), Mixer layers (i.e., MLP-Mixer~\citep{tolstikhin2021mlp}, ConvMixer~\citep{trockman2023patches}, and M2-ViT~\citep{fu2023monarch}), Transformer-based (i.e., ViT~\citep{dosovitskiy2021ViT}, DeiT-S~\citep{touvron2021training}, and Swin~\citep{liu2021swin}), and finally SSM-based baselines (i.e., VMamba~\citep{liu2024vmamba} and ViM~\citep{zhu2024ViM}). ViM2-T, ViM2-S, and ViM2-B achieve 82.7\%, 83.7\%, and 83.9\% classification accuracy, respectively. These results show that with similar scale and size, ViM2 surpasses all well-established vision models, except EffNet~\citep{Tan2019EfficientNet}. More specifically, it surpasses ViM-T by 6.6\%, ViM-S by 2.2\%, Vmamba-T by 0.5\%, Swin-T by 1.4\%, MLP-Mixer-B/16 by 6.5\%, and ResNet-50 by 6.5\%. This pattern persists across different scales: i.e., ViM2-B surpasses VMamba-B by 0.7\%, Swin-B by 0.4\%, DeiT-B by 0.8\%, ResNet-152 by 5.6\%, and EffNet-B5 by 0.3\%.

Finally, comparing ViM2-MLP with standard ViM2, results show that despite using $100\%$ more parameters, ViM2-T surpasses ViM2-MLP by 3.6\%. This shows the significance of the selective channel mixer module as replacing it with a simple MLP damages the performance significantly.

\begin{figure*}
\hspace*{-1.5ex}
    \begin{minipage}{0.5\textwidth}
        \centering
        \includegraphics[width=0.7\linewidth]{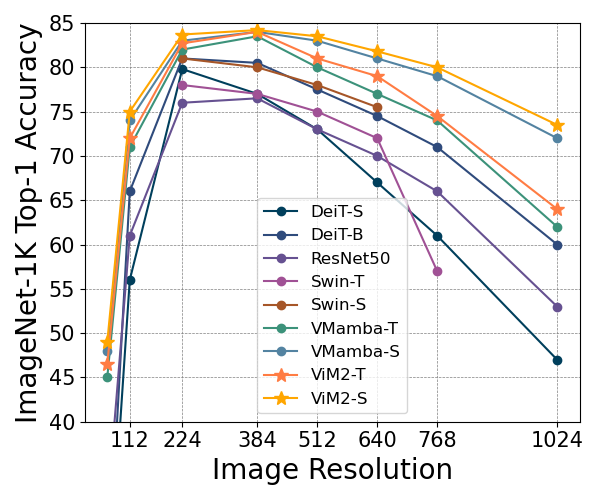}
        \caption{A comparison of input scaling evaluation for ViM2 and baselines. All models have trained with 224 × 224 inputs.}\label{fig:resolution}
    \end{minipage}\hspace{2ex}
    \begin{minipage}{0.5\textwidth}
        \centering
        \includegraphics[width=0.7\textwidth]{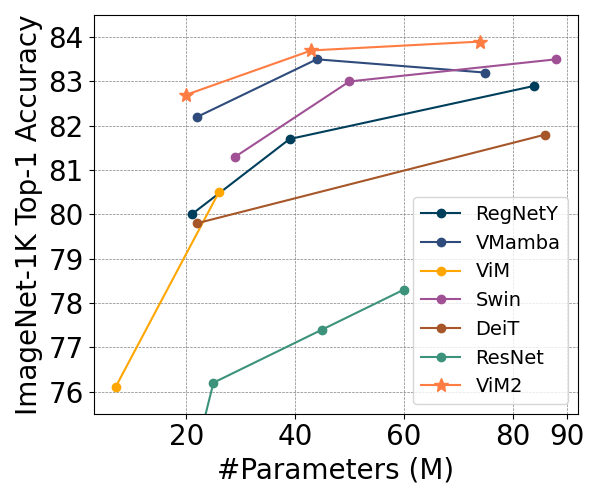}
        \caption{The effect of number of parameters and ViM2 performance comparison with baselines on ImageNet-1K.}\label{fig:parameters}
    \end{minipage}
\end{figure*}

\head{Input Scaling Evaluation and \#Parameters}
We further evaluate ViM2 on input scaling and compare it with baselines. We follow \citet{liu2024vmamba} and assess the inference performance of ViM2, which is trained with a $224 \times 224$ input size. The results are reported in \autoref{fig:resolution}. ViM2 and VMamba are the only models that show improvement when we increase image resolution from $224$ to $384$. Compared to other models, ViM2 shows the most stable performance and less performance drop when using image resolution of $1024$. We further report the performance of models on ImageNet-1K with respect to their number of parameters. The results are reported in \autoref{fig:parameters}. ViM2 shows the best performance with less number of parameters. There are two main reasons for the superior performance of ViM2. First, its hierarchical architecture allow understanding the image at different level of granularity. Second, its selective channel mixer block as well as its weighted averaging module, which allow direct access to early features, enhance the information flow, resulting in more efficient models.

\subsection{Semantic Segmentation}
\head{Setup}
In this experiment, we evaluate the performance of ViM2 model on ADE20K dataset~\citep{zhou2019semantic} and compare it with state-of-the-art baselines. Following \citet{liu2024vmamba}, we construct a UperHead~\citep{xiao2018unified} on top of the pre-trained model. We use crop size of $512 \times 512$. While again using AdamW optimizer~\citep{loshchilov2018decoupled}, we set the learning rate as $0.00006$, and use a batch size of $16$ in fine-tuning.

\head{Results}
The results of the performance comparison of ViM2 with baselines are reported in \autoref{tab:ade20k}. Results show that with similar scale and size, ViM2 outperforms all the baselines: i.e., ViM2-T surpasses VMamba-T with 1.3, ViM-S with 3.7, Swin-T with 4.2, and ResNet-50 with 6.5 mIoU (single-scale) improvement. The similar trend can be seen for ViM-S model as well as mIoU (multi-scale) measure.

\begin{table}[t]
\centering
\caption{Semantic segmentation results on ADE20K using UperNet~\citep{upernet}$^\dagger$.
}
\resizebox{.90\linewidth}{!}
{
\begin{tabular}{lcccc}
\toprule
Method & Crop size & mIoU (Single-scale) & mIoU (Multi-scale) & \#Parameters (M)  \\
\midrule
ResNet-50~\citep{He2016resnet} & $512^{2}$ & 42.1 & 42.8 & 67   \\
ResNet-101~\citep{He2016resnet} & $512^{2}$ & 42.9 & 44.0 & 85  \\
\midrule
DeiT-S + MLN~\citep{touvron2021training} & $512^{2}$ & 43.8 & 45.1 & 58  \\
DeiT-B + MLN~\citep{touvron2021training} & $512^{2}$ & 45.5 & 47.2 & 144 \\
\midrule
Swin-T~\citep{liu2021swin} & $512^{2}$ & 44.4 & 45.8 & 60  \\
Swin-S~\citep{liu2021swin} & $512^{2}$ & 47.6 & 49.5 & 81 \\
Swin-B~\citep{liu2021swin} & $512^{2}$ & 48.1 & 49.7 & 121 \\
\midrule
ConvNeXt-T~\citep{liu2022convnet}& $512^{2}$ & 46.0 & 46.7 & 60  \\
ConvNeXt-S~\citep{liu2022convnet} & $512^{2}$ & 48.7 & 49.6 & 82 \\
ConvNeXt-B~\citep{liu2022convnet} & $512^{2}$ & 49.1 & 49.9 & 122 \\
\midrule
ViM-T~\citep{zhu2024ViM} & $512^{2}$ & 41.0 & - & 13  \\
ViM-S~\citep{zhu2024ViM} & $512^{2}$ & 44.9 & - & 46 \\
\midrule
VMamba-T~\citep{liu2024vmamba} & $512^{2}$ & 47.3 & 48.3 & 55  \\
VMamba-S~\citep{liu2024vmamba} & $512^{2}$ & 49.5 & 50.5 & 76  \\
VMamba-B~\citep{liu2024vmamba} & $512^{2}$ & 50.0 & 51.3 & 110  \\
\midrule
\rowcolor{myblue}
ViM2-T & $512^{2}$ & 48.6 & 49.9 & 51  \\
\rowcolor{myblue}
ViM2-S & $512^{2}$ & 50.2 & 51.4 & 75  \\
\bottomrule
\multicolumn{5}{l}{$^\dagger$ Reported results are preliminary results and might be changed in next versions.}
\end{tabular}
}
\label{tab:ade20k}

\end{table}

\subsection{Object Detection on COCO}
\head{Setup}
In this experiment, we evaluate and compare the performance of ViM2 with baselines on object detection using the MSCOCO 2017 dataset~\citep{lin2014microsoft}. Following the experimental setup of \citet{liu2024vmamba}, we use the training framework on the mmdetection library~\citep{chen2019mmdetection}. We fine-tune the pre-trained classification models on ImageNet-1K for both 12 and 36 epochs. Again, we use AdamW optimizer with learning rate of 0.0001.

\head{Results}
The result of the performance comparison of ViM2 and baselines in object detection are reported in \autoref{tab:coco}. We found that the ability of S6 block in capturing long-range contexts results in the superior performance of ViM2 as it enables ViM2 to capture large object that other transformer-based methods fails to capture. ViM2-S achieves on par performance with VMamba-B while having almost 30\% (34 M) less parameters. Again, with similar scale and size, ViM2 surpasses baselines performance including Swin-B, ViM-T, ViT, and ResNet-101.

\begin{table}[t]
\centering
\caption{{Object detection and instance segmentation results on COCO dataset$^\dagger$}. Here, $AP^{b}$ and $AP^{m}$ denote box AP and mask AP, respectively.
}
\resizebox{.90\linewidth}{!}
{
\begin{tabular}{l ccc ccc c}
\toprule
Method & AP$^\text{b}$ & AP$^\text{b}_\text{50}$ & AP$^\text{b}_\text{75}$ & AP$^\text{m}$ & AP$^\text{m}_\text{50}$ & AP$^\text{m}_\text{75}$ & \#Parameters (M)  \\
\midrule
ResNet-50~\citep{He2016resnet} & 38.2 & 58.8 & 41.4 & 34.7 & 55.7 & 37.2 & 44 \\
ResNet-101~\citep{He2016resnet} & 38.2 & 58.8 & 41.4 & 34.7 & 55.7 & 37.2 & 63 \\
\midrule
Swin-T~\citep{liu2021swin} & 42.7 & 65.2 & 46.8 & 39.3 & 62.2 & 42.2 & 48 \\
Swin-S~\citep{liu2021swin} & 44.8 & 66.6 & 48.9 & 40.9 & 63.2 & 44.2 & 69 \\
Swin-B~\citep{liu2021swin} & 46.9 & - & - & 42.3 & - & - & 107 \\
\midrule
ConvNeXt-T~\citep{liu2022convnet} & 44.2 & 66.6 & 48.3 & 40.1 & 63.3 & 42.8 & 48 \\
ConvNeXt-S~\citep{liu2022convnet} & 45.4 & 67.9 & 50.0 & 41.8 & 65.2 & 45.1 & 70 \\
ConvNeXt-B~\citep{liu2022convnet} & 47.0 & 69.4 & 51.7 & 42.7 & 66.3 & 46.0 & 108 \\
\midrule
PVTv2-B2~\citep{wang2022pvt} & 45.3 & 67.1 & 49.6 & 41.2 & 64.2 & 44.4 & 45 \\
PVTv2-B3~\citep{wang2022pvt} & 47.0 & 68.1 & 51.7 & 42.5 & 65.7 & 45.7 & 65 \\
PVTv2-B5~\citep{wang2022pvt} & 47.4 & 68.6 & 51.9 & 42.5 & 65.7 & 46.0 & 102 \\
\midrule
ViT-Adapter-S~\citep{dosovitskiy2021ViT} & 44.7 & 65.8 & 48.3 & 39.9 & 62.5 & 42.8 & 48 \\
ViT-Adapter-B~\citep{dosovitskiy2021ViT} & 47.0 & 68.2 & 51.4 & 41.8 & 65.1 & 44.9 & 102 \\
\midrule 
ViM-T~\citep{zhu2024ViM} &45.7 & 63.9   & 49.6 & 26.1 & 49.0 & 63.2 & -$^*$ \\
\midrule
VMamba-T~\citep{liu2024vmamba} & 46.5 & 68.5 & 50.7 & 42.1 & 65.5 & 45.3 & 42\\
VMamba-S~\citep{liu2024vmamba} & 48.2 & 69.7 & 52.5 & 43.0 & 66.6 & 46.4 & 64 \\ 
VMamba-B~\citep{liu2024vmamba} & 48.5 & 69.6 & 53.0 & 43.1 & 67.0 & 46.4 & 96 \\
\midrule
\rowcolor{myblue}
ViM2-T & 47.1 & 68.7 & 50.9 & 42.4 & 65.6 & 45.5 & 39 \\
\rowcolor{myblue}
ViM2-S & 48.5 & 69.9 & 52.8 & 43.1 & 66.8 & 46.5 & 62\\
\bottomrule
\multicolumn{8}{l}{$^*$ The number of parameters is not reported~\citep{zhu2024ViM}.}\\
\multicolumn{8}{l}{$^\dagger$ Reported results are preliminary results and might be changed in next versions.}
\end{tabular}
}
\label{tab:coco}
\end{table}

\subsection{Multivariate Long-term Forecasting}
\head{Setup}
We perform experiments on 8 publicly available benchmark datasets of real-world multivariate time series, commonly used in studies on long-term forecasting~\citep{ilbert2024unlocking, chen2023tsmixer, nie2023a}. We compare the performance of TSM2 with the state-of-the-art models on multivariate time series forecasting: i.e., SAMFormer~\citep{ilbert2024unlocking}, simple cross-variate Transformer~\citep{ilbert2024unlocking}, TSMixer~\citep{chen2023tsmixer}, Informer~\citep{zhou2021informer}, Autoformer~\citep{wu2021autoformer}, FEDFormer~\citep{zhou2022fedformer}, Pyraformer~\citep{liu2021pyraformer}, and LogTransformer~\citep{li2019enhancing}. All time series are segmented with input length $\texttt{L} = 512$, prediction horizons $\texttt{H}\in \{96, 192, 336, 720\}$, and a stride of 1.

\head{Results}
We compare the MSE results of TSM2 with baselines' in \autoref{tab:TSM2_results}. TSM2 outperforms all the baselines in most cases (26 out of 32 cases as the best model and 5 as the second best model).  The reason for the superior performance of TSM2 is two folds:\circled{1} While some baselines have focused more on cross-time dependencies~\citep{zhou2021informer, zhou2022fedformer}, some others are designed to focus more on cross-variate dependencies~\citep{ilbert2024unlocking}. This choice, however, depends on the nature of the dataset. TSM2 using weighted averaging module can learn whether it needs to focus more on cross-variate or cross-time dependencies in a data-driven manner. \circled{2} The dual selection mechanism in MambaMixer not only allows TSM2 to capture long-range dependencies but it also uses data-dependent weights, making model more generalizable; while some baselines like TSMixer~\citep{chen2023tsmixer} uses data-independent simple MLPs to fuse information, lacking the selection ability.


\begin{table*}
\centering
\caption{Performance comparison between our {MambaMixer} and baselines for multivariate long-term forecasting with different horizons \texttt{H}. Results of baselines are obtained from~\citet{ilbert2024unlocking}. 
We display the MSE of methods. Best results are bolded and highlighted in {blue}, second best are highlighted in {gray}$^\dagger$.}
\label{tab:TSM2_results}
\hspace*{-5ex}
\resizebox{1.1\linewidth}{!}{
\begin{tabular}{ccccccccccc}
\toprule%
\multicolumn{1}{c}{\multirow{2}{*}{Dataset}} & \multicolumn{1}{c}{\multirow{2}{*}{\texttt{H}}} & \multicolumn{1}{c}{TSMambaMixer} & \multicolumn{1}{c}{SAMFormer}  & \multicolumn{1}{c}{Transformer} & \multicolumn{1}{c}{TSMixer} & \multicolumn{1}{c}{Informer} & \multicolumn{1}{c}{Autoformer} & \multicolumn{1}{c}{FEDFormer} & \multicolumn{1}{c}{Pyraformer} & \multicolumn{1}{c}{LogTransformer}\\ 
 \multicolumn{1}{c}{} & \multicolumn{1}{c}{} & (\textcolor{c2}{Ours}) & \citep{ilbert2024unlocking} & \citep{ilbert2024unlocking} & \citep{chen2023tsmixer} & \citep{zhou2021informer} & \citep{wu2021autoformer} & \citep{zhou2022fedformer} & \citep{liu2021pyraformer} & \citep{li2019enhancing}\\
\midrule
\multirow{4}{*}{\rotatebox[origin=c]{90}{ETTh1}} & $96$ & \cellcolor{myblue}$\mathbf{0.375}$ & ${0.381}$  & $0.509$ & $0.398$ & $0.941$ & $0.435$ & \cellcolor{mygray}${0.376}$ & 0$.664$ & $0.878$ \\
& $192$ & \cellcolor{myblue}$\mathbf{0.398}$ & \cellcolor{mygray}${0.409}$ & $0.535$ & $0.426$ & $1.007$ &$ 0.456$ & $0.423$ & $0.790$  & $1.037$\\
& $336$ & \cellcolor{myblue}$\mathbf{0.419}$ & \cellcolor{mygray}${0.423}$  & $0.570$ & $0.435$ & $1.038$ & $0.486$ & $0.444$ & $0.891$ & $1.238$\\ 
& $720$ & \cellcolor{myblue}$\mathbf{0.422}$ & \cellcolor{mygray}${0.427}$ & $0.601$ & $0.498$ & $1.144$ & $0.515$ & $0.469$ & $0.963$ & $1.135$\\
\midrule
\multirow{4}{*}{\rotatebox[origin=c]{90}{ETTh2}} & $96$ & \cellcolor{myblue}$\mathbf{0.253}$ & \cellcolor{mygray}${0.295}$  & $0.396$ & $0.308$ & $1.549$ & $0.332$ & $0.332$ & $0.645$ & $2.116$\\
& $192$ & \cellcolor{myblue}$\mathbf{0.334}$ & \cellcolor{mygray}${0.340}$ & $0.413$ & $0.352$ & $3.792$ & $0.426$ & $0.407$ &$ 0.788$ & $4.315$\\
& $336$ & \cellcolor{myblue}$\mathbf{0.347}$ & \cellcolor{mygray}${0.350}$  & $0.414$ & $0.360$ & $4.215$ & $0.477 $& $0.400$ & $0.907$ & $1.124$\\ 
& $720$ & \cellcolor{mygray}${0.401}$ & \cellcolor{myblue}$\mathbf{0.391}$  & $0.424$ & $0.409$ & $3.656 $& $0.453$ & $0.412$ & $0.963$ & $3.188$\\
\midrule
\multirow{4}{*}{\rotatebox[origin=c]{90}{ETTm1}} & $96$ & \cellcolor{myblue}$\mathbf{0.322}$ & $0.329$ & $0.384$ & $0.336$ & $0.626$ & $0.510$ & \cellcolor{mygray}${0.326}$ & $0.543$ & $0.600$\\
& $192$ & \cellcolor{myblue}$\mathbf{0.349}$ & \cellcolor{mygray}${0.353}$ & $0.400$ & $0.362$ & $0.725$ & $0.514$ & $0.365$ & $0.557$ & $0.837$\\
& $336$ & \cellcolor{myblue}$\mathbf{0.366}$ & \cellcolor{mygray}${0.382}$ & $0.461$ & $0.391$ & $1.005$ & $0.510$ & $0.392$ &$ 0.754$ & $1.124$\\ 
& $720$ & \cellcolor{myblue}$\mathbf{0.407}$ & \cellcolor{mygray}${0.429}$  & $0.463$ & $0.450$ & $1.133$ & $0.527$ & $0.446$ & $0.908$ & $1.153$\\
\midrule
\multirow{4}{*}{\rotatebox[origin=c]{90}{ETTm2}} & $96$ & \cellcolor{myblue}$\mathbf{0.173}$ & ${0.181}$ & $0.200$ & $0.211$ & $0.355$ & $0.205$ & \cellcolor{mygray}${0.180}$ & $0.435$ & $0.768$ \\
& $192$ & \cellcolor{myblue}$\mathbf{0.230}$ & \cellcolor{mygray}${0.233}$ & $0.273$ & $0.252$ & $0.595$ & $0.278$ & $0.252$ & $0.730$ &$ 0.989$\\
& $336$ & \cellcolor{myblue}$\mathbf{0.279}$ & \cellcolor{mygray}${0.285}$ & $0.310$ & $0.303$ & $1.270$ & $0.343$ & $0.324$ & $1.201$ & $1.334$\\ 
& $720$ & \cellcolor{mygray}${0.388}$ & \cellcolor{myblue}$\mathbf{0.375}$  & $0.426$ & $0.390$ & $3.001$ & $0.414$ & $0.410$ & $3.625$ & $3.048$\\
\midrule
\multirow{4}{*}{\rotatebox[origin=c]{90}{\text{\small Electricity}}} & $96$ & \cellcolor{myblue}$\mathbf{0.142}$ & \cellcolor{mygray}${0.155}$ & $0.182$ & $0.173$ & $0.304$ & $0.196$ & $0.186$ & $0.386$ & $0.258$\\
& $192$ & \cellcolor{myblue}$\mathbf{0.153}$ & \cellcolor{mygray}${0.168}$ & $0.202$ & $0.204$ & $0.327$ & $0.211$ & $0.197$ & $0.386$ & $0.266$\\
& $336$ & \cellcolor{myblue}$\mathbf{0.175}$ & \cellcolor{mygray}${0.183}$  & $0.212$ & $0.217$ & $0.333$ & $0.214$ & $0.213$ & $0.378$ & $0.280$\\ 
& $720$ & \cellcolor{myblue}$\mathbf{0.209}$ & \cellcolor{mygray}${0.219}$  & $0.238$ & $0.242$ & $0.351$ & $0.236$ & $0.233$ & $0.376$ & $0.283$\\
\midrule
\multirow{4}{*}{\rotatebox[origin=c]{90}{Exchange}} 
& $96$ & ${0.163}$ & \cellcolor{mygray}${0.161}$ & $0.292$ & $0.343$ & $0.847$ & $0.197$ & \cellcolor{myblue}$\mathbf{0.139}$ &-&$0.968$\\
& $192$ & \cellcolor{myblue}$\mathbf{0.229}$ & \cellcolor{mygray}${0.246}$  & $0.372$ & $0.342$ &$1.204$ & $0.300$ & $0.256$&-&$1.040$\\
& $336$ & \cellcolor{mygray}${0.383}$ & \cellcolor{myblue}$\mathbf{0.368}$ & $0.494$ & $0.484$ &$1.672$ & $0.509$ & $0.426$&-&$1.659$\\ 
& $720$ & \cellcolor{myblue}$\mathbf{0.999}$ & \cellcolor{mygray}${1.003}$  & $1.323$ & $1.204$ &$2.478$ & $1.447$ & $1.090$&-&$1.941$\\
\midrule
\multirow{4}{*}{\rotatebox[origin=c]{90}{Traffic}} 
& $96$ & \cellcolor{myblue}$\mathbf{0.396}$ & \cellcolor{mygray}${0.407}$ & $0.420$ & $0.409$ & $0.733$ & $0.597$ & $0.576$ & $2.085$ & $0.684$\\
& $192$ & \cellcolor{myblue}$\mathbf{0.408}$ & \cellcolor{mygray}${0.415}$  & $0.441$ & $0.637$ & $0.777$ & $0.607$ & $0.610$ & $0.867$ & $0.685$\\
& $336$ & \cellcolor{mygray}${0.427}$ & \cellcolor{myblue}$\mathbf{0.421}$ & $0.501$ & $0.747$ & $0.776$ & $0.623$ & $0.608$ & $0.869$ & $0.734$\\ 
& $720$ & \cellcolor{myblue}$\mathbf{0.449}$ & \cellcolor{mygray}${0.456}$  & $0.468$ & $0.688$ & $0.827$ & $0.639$ & $0.621$ & $0.881$ & $0.717$\\
\midrule
\multirow{4}{*}{\rotatebox[origin=c]{90}{Weather}} 
& $96$ & \cellcolor{myblue}$\mathbf{0.161}$ & \cellcolor{mygray}${0.197}$  & $0.227$ & $0.214$ & $0.354$ & $0.249$ & $0.238$ & $0.896$ & $0.458$\\
& $192$ & \cellcolor{myblue}$\mathbf{0.208}$ & ${0.235}$  & $0.256$ & \cellcolor{mygray}$0.231$ & $0.419$ & $0.325$ & $0.275$ & $0.622$ & $0.658$\\
& $336$ & \cellcolor{myblue}$\mathbf{0.252}$ & \cellcolor{mygray}${0.276}$ & $0.278$ & $0.279$ & $0.583$ & $0.351$ & $0.339$ & $0.739$ & $0.797$\\ 
& $720$ & \cellcolor{mygray}${0.337}$ & \cellcolor{myblue}$\mathbf{0.334}$  & $0.353$ & $0.343$ & $0.916$ & $0.415$ & $0.389$& $1.004$ &$ 0.869$\\
\bottomrule
\multicolumn{10}{l}{$^\dagger$ Reported results are preliminary results and might be changed in the next versions.}
\end{tabular}
}
\end{table*}

\head{Forecasting with Auxiliary Information}
\citet{chen2023tsmixer} discussed the importance of auxiliary information in time series forecasting. To show the ability of our TSM2 in using auxiliary features, we follow \citet{chen2023tsmixer} and perform an experiment on M5 dataset~\citep{makridakis2022m5}. We use two additional baselines that are capable of using auxiliary features: i.e., DeepAR~\citep{salinas2020deepar} and TFT~\citep{lim2021temporal}. The results are reported in \autoref{tab:auxiliary_result}. TSM2 significantly outperforms baselines with 7.43\% improvement over the second best model, i.e., TSMixer~\citep{chen2023tsmixer}.

\begin{table}[ht]
    \centering
        \caption{Evaluation of TSM2 and baselines on M5 dataset with auxiliary information.}
    \resizebox{0.8\linewidth}{!}{    
    \begin{tabular}{l c c c c}
    \toprule
         {Measure}& {TSMambaMixer} & TSMixer & DeepAR & TFT \\
         & (\textcolor{c2}{Ours}) & \citep{chen2023tsmixer} & \citep{salinas2020deepar} & \citep{lim2021temporal}\\
         \midrule
         {Test WRMSSE} & \cellcolor{myblue} \textbf{0.591} & \cellcolor{mygray}0.640 & 0.789 & 0.670 \\
         {Val WRMSSE} & \cellcolor{myblue} \textbf{0.527} & \cellcolor{mygray}0.568 & 0.611 & 0.579 \\
    \toprule
    \end{tabular}
    }
    \label{tab:auxiliary_result}
\end{table}

\subsection{Significance of Dual Selection}
We further evaluate the significance of dual selection and the architectural design of MambaMixer block in TSM2. To this end, we use two variants of TSM2: \circled{1} TSM2-MLP replaces the selective channel mixer block with MLP. \circled{2} Mamba + Linear Time: use unidirectional Mamba~\citep{gu2023mamba} as the channel mixer and linear layer as the token mixer (similar to TSMixer architecture~\citep{chen2023tsmixer}). The results are reported in \autoref{tab:ablation-study}. Compared to TSM2-MLP, the superior performance of TSM2 shows the significance of selective channel mixer block in TSM2. Compared to the second baseline, the superior performance of TSM2 shows the importance of bidirectionality in the design of selective channel mixer as well as the significance of the selective token mixer block.

\begin{table}[ht]
    \centering
        \caption{Ablation study on the architecture of TSM2 on ETTh1, ETTm1, Exchange, and Electricity datasets.}
    \resizebox{0.7\linewidth}{!}{    
    \begin{tabular}{l c c c c}
    \toprule
         {Model}& {ETTh1} & ETTm1 & Exchange & Electricity \\
         \midrule
         {TSM2} & \cellcolor{myblue}\textbf{0.375} & \cellcolor{myblue}\textbf{0.322} & \cellcolor{myblue}\textbf{0.163} & \cellcolor{myblue}\textbf{0.142} \\
         {TSM2-MLP} & 0.386 & 0.339 & 0.197 & 0.173 \\
         {Mamba + Linear Time} & 0.388 & 0.334 & 0.220 & 0.151\\
    \toprule
    \end{tabular}
    }
    \label{tab:ablation-study}
\end{table}

\section{Conclusion}
We present MambaMixer, an efficient selective state space model with dual token and channel selection. MambaMixer uses selective state space models (S6 blocks) in both token and channel directions, enabling the model to effectively and selectively fuse information across both of these dimensions. To enhance the information flow and capturing the complex dynamics of features, MambaMixer uses a learnable weighted averaging mechanism on early features, which allows each block to directly access to early features. As proof of concept, we further present ViM2 and TSM2 models based on MambaMixer block for vision and time series forecasting tasks. Our experimental evaluations show that In ImageNet classification, object detection, and semantic segmentation tasks, ViM2 achieves competitive performance with well-established vision models, i.e., ViT, MLP-Mixer, ConvMixer, and outperforms SSM-based vision models, i.e., ViM and VMamba. In time series forecasting, TSM2 outperforms all the baselines in most datasets and achieve state-of-the-art performance while demonstrating significantly improved computational cost.


\newpage
\bibliographystyle{icml2024}
\bibliography{main}

\newpage
\appendix

\section{Architectural Overview}

\begin{table}[!ht]
\begin{center}
\caption{
Architectural overview of the ViM2 in different scales.
}\label{tab:scales}
\resizebox{.95\linewidth}{!}
{
\begin{tabular}{c c c c c c}
\toprule
layer name & output size &  & Tiny & Small & Base  \\
\midrule
Stem & 112$\times$112 & & conv 4$\times$4, 96, stride 4 & conv 4$\times$4, 96, stride 4 & conv 4$\times$4, 128, stride 4\\
\midrule
\multirow{14}{*}{Stage 1} & \multirow{14}{*}{56$\times$56} & \multirow{7}{*}{\rotatebox[origin=c]{90}{Token Mixer}} & &  \\ 
&  & & \blocka{96}{2}{16}  & \blocka{96}{2}{16} & \blocka{128}{2}{16} \\
&  & &  &  & \\
&  & &  &  & \\
&  & &  &  & \\
&  & &  &  & \\
 &  &   \multirow{7}{*}{\rotatebox[origin=c]{90}{Channel Mixer}} & &  \\ 
&  &  & \blockb{192}{1}  & \blockb{192}{1} & \blockb{256}{1} \\
&  & & &  & \\
&  & & &  & \\
&  & & &  & \\
&  & & &  & \\
& & & & &\\
\midrule
\multirow{14}{*}{Stage 2} & \multirow{14}{*}{28$\times$28} & \multirow{7}{*}{\rotatebox[origin=c]{90}{Token Mixer}}&  &  \\ 
&  & & \blocka{192}{2}{16}  & \blocka{192}{2}{16} & \blocka{256}{2}{16} \\
&  & & &  & \\
&  & & &  & \\
&  & &  &  & \\
&  & &  &  & \\
&  & &  &  &\\
&  & \multirow{7}{*}{\rotatebox[origin=c]{90}{Channel Mixer}} & &  &  \\ 
&  &  & \blockb{384}{1}  & \blockb{384}{1} & \blockb{512}{1} \\
&  & & &  & \\
&  & & &  & \\
&  & & &  & \\
&  & & &  & \\
&  & & &  &\\
\midrule
\multirow{14}{*}{Stage 3} & \multirow{14}{*}{14$\times$14} & \multirow{7}{*}{\rotatebox[origin=c]{90}{Token Mixer}} & &  \\ 
&  & & \blocka{384}{6}{16}  & \blocka{384}{18}{16} & \blocka{512}{18}{16} \\
&  & &  &  & \\
&  & &  &  & \\
&  & &  &  & \\
&  &  & &  & \\
 & & \multirow{7}{*}{\rotatebox[origin=c]{90}{Channel Mixer}} & &  \\ 
&  &  & \blockb{768}{3}  & \blockb{768}{9} & \blockb{1024}{9} \\
&  & &  &  & \\
&  & &  &  & \\
&  & &  &  & \\
&  &  & &  & \\
& & & & &\\
\midrule
\multirow{15}{*}{Stage 4} & \multirow{15}{*}{7$\times$7} & \multirow{7}{*}{\rotatebox[origin=c]{90}{Token Mixer}} & & \\ 
&  & & \blocka{768}{2}{16}  & \blocka{768}{2}{16} & \blocka{1024}{2}{16} \\
&  & & &  & \\
&  & & &  & \\
&  & & &  & \\
&  & & &  & \\
 &  & & & \\ 
  &  & \multirow{7}{*}{\rotatebox[origin=c]{90}{Channel Mixer}}& & \\ 
&  &  & \blockb{768}{1}  & \blockb{768}{1} & \blockb{1024}{1} \\
&  & & &  & \\
&  & & &  & \\
&  & & &  & \\
&  & & &  & \\
& &  & & &\\
\midrule
\text{Output}& 1$\times$1  & & \multicolumn{3}{c}{Average Pooling, 1000D FC, \texttt{Softmax}} \\
\toprule
\end{tabular}
}
\end{center}
\end{table}

\section{Possible Architecture Design of Selective Channel and Token Mixer Blocks}\label{app:arch}

\begin{figure}[!ht]
    \centering
    \includegraphics[width=\linewidth]{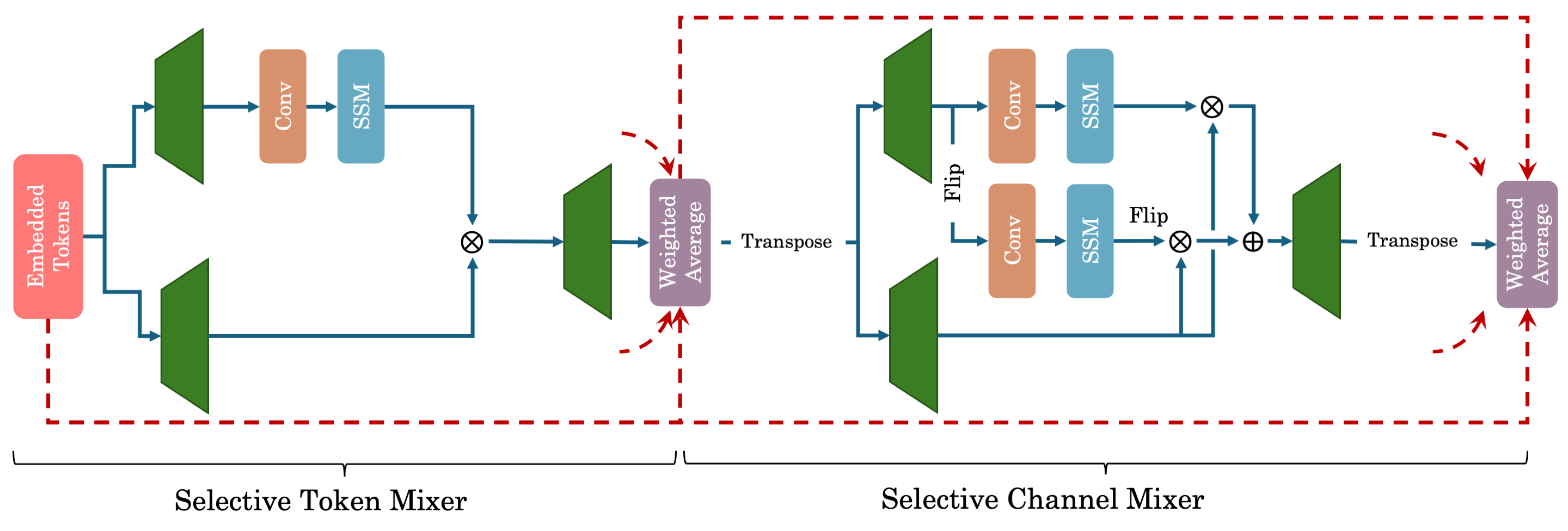}
    \caption{Architecture design of MambaMixer (unidirectional token mixer).}
    \label{fig:M2Arch1-2}
\end{figure}

\begin{figure}[!ht]
    \centering
    \includegraphics[width=\linewidth]{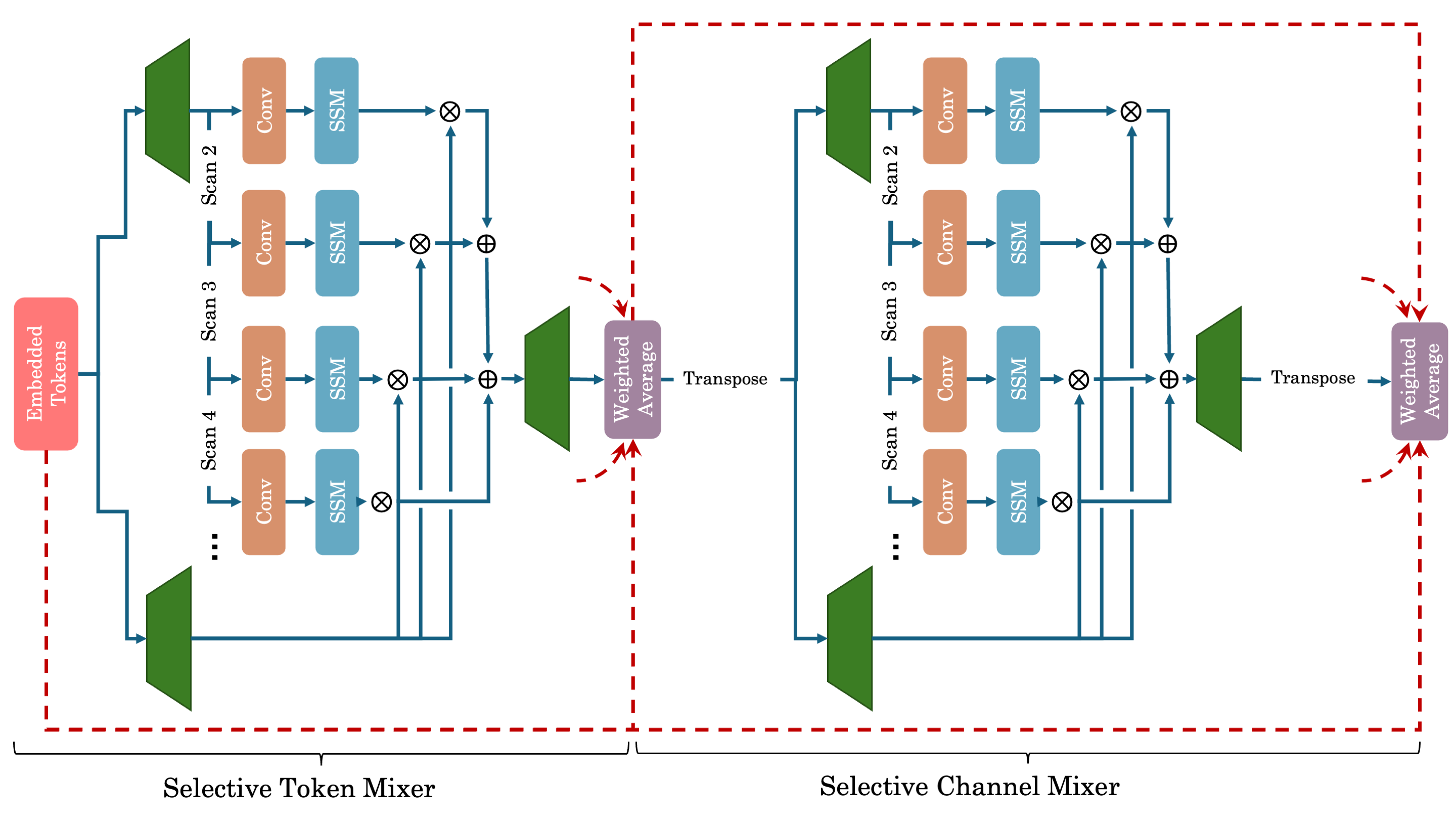}
    \caption{Architecture design of MambaMixer (with $n$ scans).}
    \label{fig:M2Arch-n-n}
\end{figure}

\begin{figure}[!ht]
    \centering
    \includegraphics[width=\linewidth]{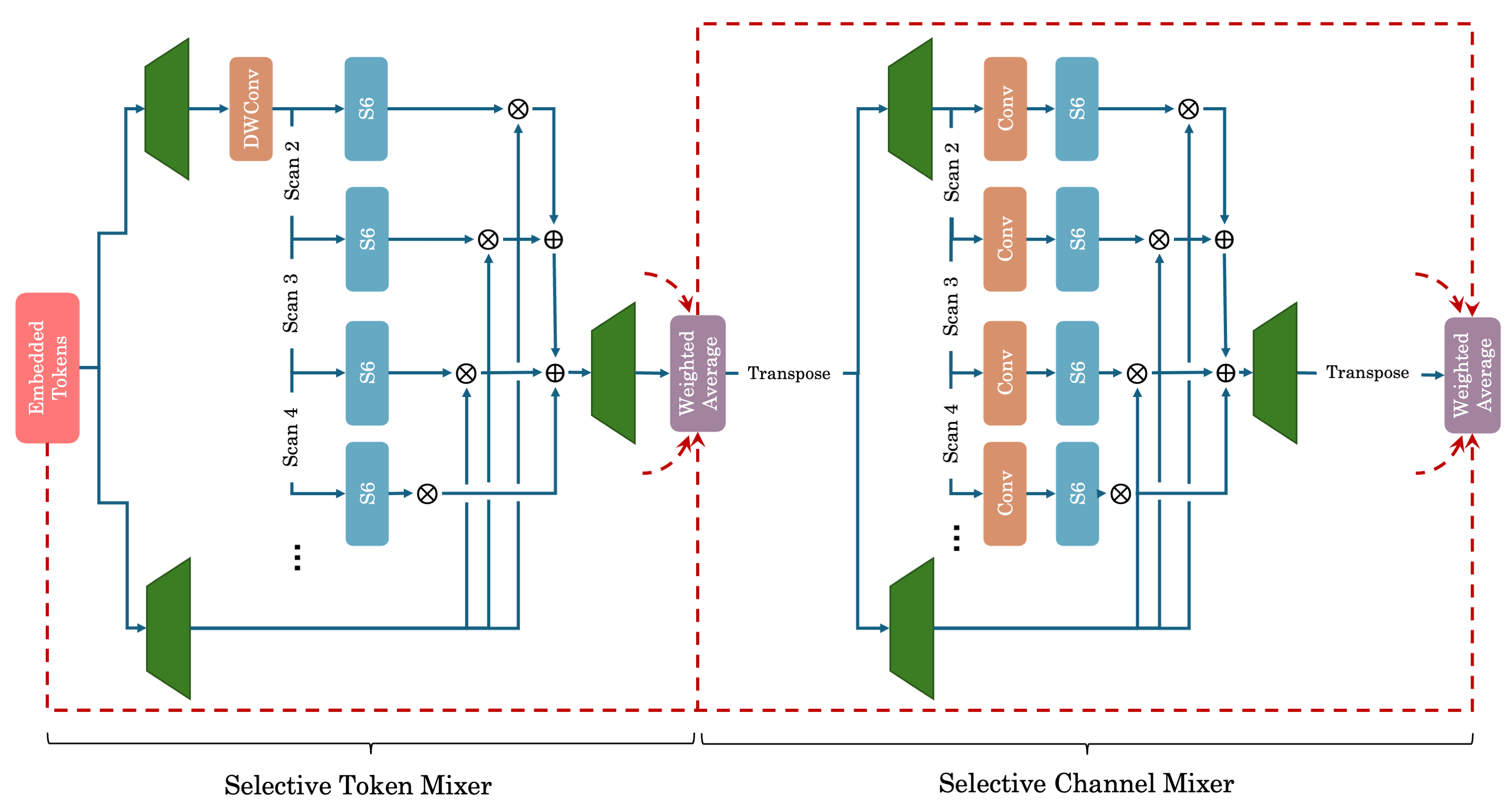}
    \caption{Architecture design of MambaMixer used in ViM2 with depth-wise Conv.}
    \label{fig:ViM2-n-n}
\end{figure}

\end{document}